\documentclass[10pt,twocolumn,letterpaper]{article}

\usepackage{cvpr}              % To produce the CAMERA-READY version

\usepackage{array}
\usepackage{graphicx}
\usepackage{amsmath}
\usepackage{amssymb}
\usepackage{multirow}
\usepackage{booktabs}
\usepackage{url}            % simple URL typesetting
\usepackage{amsfonts}       % blackboard math symbols
\usepackage{nicefrac}       % compact symbols for 1/2, etc.
\usepackage{microtype}      % microtypography
\usepackage{colortbl}
\usepackage{pifont} 
\usepackage{listings} 
\usepackage{wrapfig}
\usepackage{enumitem} 
\usepackage{amsthm}
\usepackage{tcolorbox}
\definecolor{cvprblue}{rgb}{0.21,0.49,0.74}
\definecolor{featcol}{RGB}{153,51,255}
\definecolor{featray}{RGB}{0,0,255}
\definecolor{pos}{RGB}{0,127,255}\usepackage[pagebackref,breaklinks,colorlinks,citecolor=cvprblue]{hyperref}

\def\paperID{42} 
\def\confName{3DV\xspace}
\def\confYear{2024\xspace}

\definecolor{Gray}{gray}{0.85}
\definecolor{GrayBorder}{gray}{0.65}

\theoremstyle{plain}
\newtheorem{thm}{Theorem}[section] % reset theorem numbering for each chapter

\newcolumntype{a}{>{\columncolor{GrayBorder}}l}
\newcolumntype{b}{>{\columncolor{Gray}}c}
\newcolumntype{d}{>{\columncolor{GrayBorder}}c}

\newcommand{\method}{Zero-BEV}

\newcommand{\rgb}{^{rgb}}
\newcommand{\aux}{^{aux}}
\newcommand{\zero}{^{zero}}

\definecolor{codegreen}{rgb}{0,0.6,0}
\definecolor{codegray}{rgb}{0.5,0.5,0.5}
\definecolor{codepurple}{rgb}{0.58,0,0.82}
\definecolor{backcolour}{rgb}{0.95,0.95,0.92}

\lstdefinestyle{mystyle}{
    backgroundcolor=\color{backcolour},   
    commentstyle=\color{codegreen},
    keywordstyle=\color{magenta},
    numberstyle=\tiny\color{codegray},
    stringstyle=\color{codepurple},
    basicstyle=\ttfamily\footnotesize,
    breakatwhitespace=false,         
    breaklines=true,                 
    captionpos=b,                    
    keepspaces=true,                 
    numbers=left,                    
    numbersep=5pt,                  
    showspaces=false,                
    showstringspaces=false,
    showtabs=false,                  
    tabsize=2
}

\newcommand{\myparagraph}[1]{\textbf{#1} ---}

\title{Zero-BEV: Zero-shot Projection of Any First-Person Modality to BEV Maps}

\author{Gianluca Monaci, Leonid Antsfeld, Boris Chidlovskii, Christian Wolf \\
NAVER LABS Europe, Meylan, France\\
{\tt\small firstname.lastname@naverlabs.com}
}

\begin{document}
\maketitle

\begin{abstract}
\noindent
Bird's-eye view (BEV) maps are an important geometrically structured 
representation widely used in robotics, in particular self-driving vehicles and terrestrial robots. 
Existing algorithms either require depth information for the geometric projection, which is not always reliably available, or are trained end-to-end in a fully supervised way to map visual first-person observations to BEV representation, and are therefore restricted to the output modality they have been trained for. In contrast, we propose a new model capable of performing zero-shot projections of any modality available in a first person view to the corresponding BEV map.
This is achieved by disentangling the geometric inverse perspective projection from the modality transformation, eg. RGB to occupancy.
The method is general and we showcase experiments projecting to BEV three different modalities: semantic segmentation, motion vectors and object bounding boxes detected in first person. We experimentally show that the model outperforms competing methods, in particular the widely used baseline resorting to monocular depth estimation.
\end{abstract}

\begin{figure}[t]  \centering
    \includegraphics[width=\linewidth]{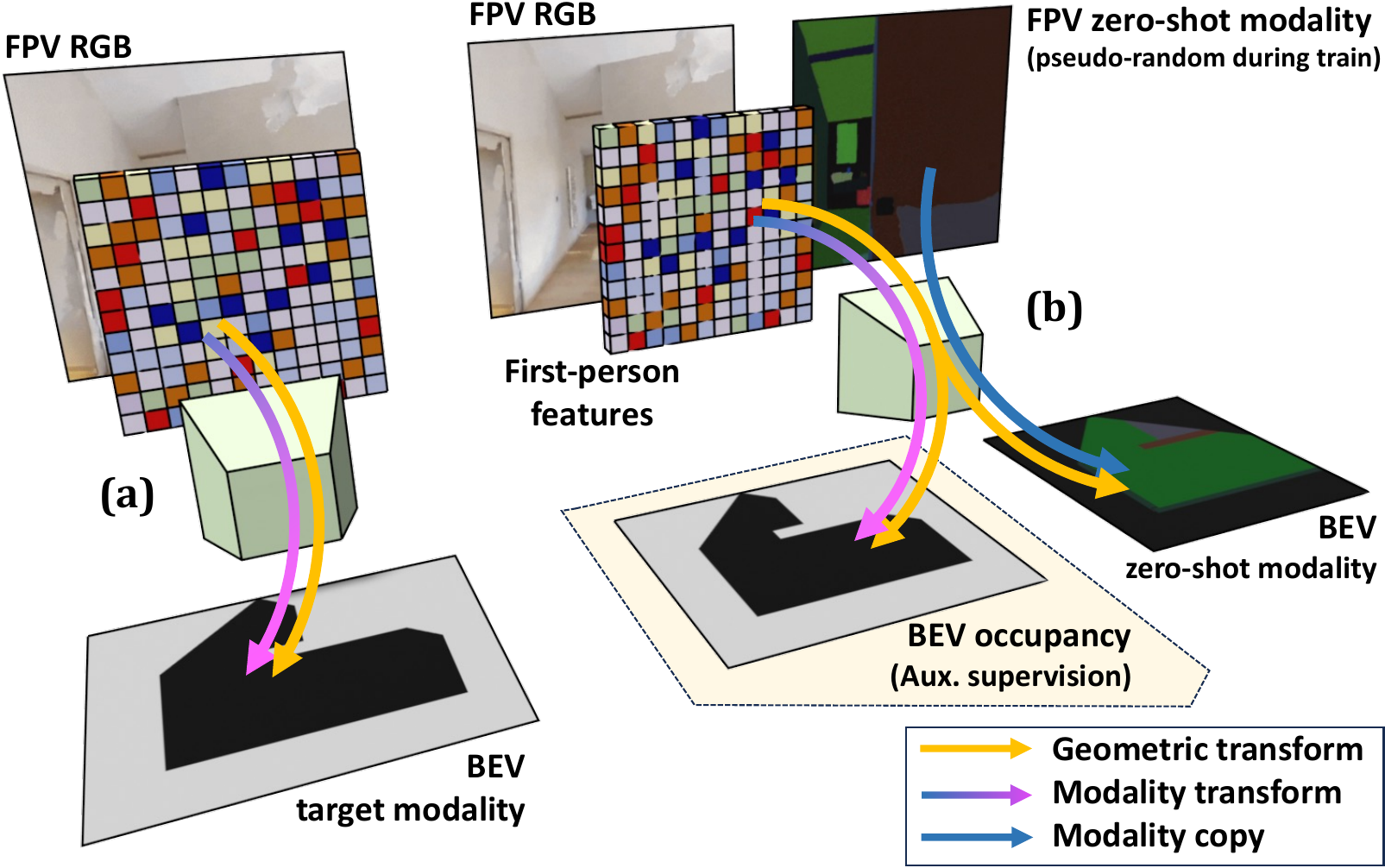}
    \caption{\label{fig:teaser}We train a model to project first-person views (FPVs) to BEV maps. (a) Existing work trains end-to-end the prediction of the target modality. (b) We disentangle two underlying transformations: \ding{192} the geometric projection from FPV to BEV \includegraphics[width=0.4cm,height=2mm]{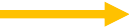}, and \ding{193} an optional modality translation seen during training \includegraphics[width=0.4cm,height=2mm]{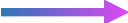}, eg. RGB to occupancy. This enables zero-shot projection of any modality unseen during training at deployment, leaving the modality unchanged \includegraphics[width=0.4cm,height=2mm]{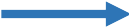} and performing the geometric transformation only.}
\end{figure}

%%%%%%%%%%%%%%%%%%%%%%%%%%%%%%%%%%%%%%%%%%%%%%%%%%%%%%%%%%%%%%%%%%%%%%%%%%%%%%%%%%%%%%%%%%%%%%%%%
%%%%%%%%%%%%%%%%%%%%%%%%%%%%%%%%%%%%%%%%%%%%%%%%%%%%%%%%%%%%%%%%%%%%%%%%%%%%%%%%%%%%%%%%%%%%%%%%%
\section{Introduction}
\label{sec:intro}
Embodied agents such as robots or autonomous cars require situation awareness and an understanding of their spatial surroundings to effectively operate in their environment. A natural, convenient and widely used representation is based on {\it bird's-eye view} (BEV) maps, which exhibit several interesting properties: they are free of the perspective distortion inherent to sensing and, as such, closer to the Euclidean space in which the agent operates; they can compactly represent a wealth of different modalities like occupancy~\cite{thrun2005probabilistic,labbe19rtabmap,chaplot2019learning}, semantics~\cite{chaplot2020object, ramakrishnan2022poni}, motion vectors~\cite{RummelhardCMCDOT2015,ErkentIROS2018} or even latent representations~\cite{henriques2018mapnet,beeching2020egomap,marza2021teaching}; they naturally allow integration of multiple first-person views (FPVs)~\cite{bartoccioni2022lara}. Finally, they can be broadly used for downstream tasks and, depending on the stored modality, can be used for direct planning with optimal path planners, and have been shown to transfer well from simulation to the real world in hybrid architectures~\cite{gervet2023navigating, chang2023goat}.

While classical approaches to BEV map reconstruction were based on geometry and probabilistic models~\cite{thrun2005probabilistic} using LiDAR~\cite{Grisetti2005ParticleFilter, RummelhardCMCDOT2015}, this work focuses on \emph{vision-based} BEV map creation~\cite{ma2023vision}. Early attempts use FPV image input and inverse projection to generate BEV views, assuming flat ground and fixed camera height~\cite{aly2008real,sengupta2012automatic}. More recent works resort to deep learning to infer BEV maps from FPV images \cite{pan2019crossview, saha2022translating, bartoccioni2022lara}, Figure \ref{fig:teaser}(a).
These learning-based methods have the remarkable property of exploiting data regularities to infer details that are \textit{not necessarily visible from first-person view}, e.g. partially occluded objects and spaces. However, they are trained fully supervised with BEV maps labeled with target classes.

This paper presents a new approach and zero-shot task that maintains the advantages of both geometry and learning-based methods, Figure \ref{fig:teaser}(b): (i) generate BEV maps using FPV RGB images and by zero-shot projection of any arbitrary modality additionally available in the FPV: occupancy, semantics, object instances, motion vectors etc. (ii) exploit regularities to infer information not visible in FPV.

Different modalities are typically readily available from FPV input through pre-trained segmenters and detectors, which are %trained to be 
agnostic of camera resolutions and intrinsics. Projecting this information to a BEV map, however, is trivial only for methods that explicitly model geometry and only if depth information is available, or can be estimated reliably% and with high quality
, which is not always the case. For the family of models based on end-to-end supervised learning, prior work requires training a mapping function separately for each projected modality, which is cumbersome and requires costly BEV annotations~\cite{meng2020weakly}. These methods also require retraining if the modality is modified, e.g. if an object class is added.

We propose a new learning approach capable of zero-shot projecting any modality available in FPV images to a BEV map, \textit{without} requiring depth input. This is achieved through a single training step, where we disentangle \ding{192} the geometric transformation between FPV to BEV and \ding{193} the modality translation, e.g. from RGB values to occupancy, semantics etc. (see Figure~\ref{fig:teaser}). We explore and evaluate two different ways to achieve disentanglement: a new data generation procedure decorrelating geometry from other scene properties, and optionally, an additional inductive bias for the cross-attention layers of a transformer-based architecture.

%%%%%%%%%%%%%%%%%%%%%%%%%%%%%%%%%%%%%%%%%%%%%%%%%%%%%%%%%%%%%%%%%%
%%%%%%%%%%%%%%%%%%%%%%%%%%%%%%%%%%%%%%%%%%%%%%%%%%%%%%%%%%%%%%%%%%
\section{Related work}
\label{sec:sota}

\noindent
\myparagraph{Geometry-based BEV} 
Early attempts used inverse perspective mapping~\cite{mallot91inverse} to project image pixels to the BEV plane assuming a flat ground and using the geometric constraint of camera intrinsic and extrinsic matrices~\cite{aly2008real,sengupta2012automatic}.
However, the flat-world assumption fails on non-flat content thus producing artifacts in the ground plane, like shadows or cones. 
More recent approaches~\cite{liu20understanding, wang21probabilistic, schulter18learning, chaplot2020object, gervet2023navigating, chang2023goat} use depth and semantic segmentation maps (required as input) to lift 2D objects in 3D and then project them into BEV. 

\myparagraph{Vision-based semantic BEV segmentation}
Vision-centric methods count on getting richer semantic information from images and rely on high-level understanding of them to reason on the scene geometry~\cite{li22delving}. 
Early methods learn end-to-end FPV to BEV mappings~\cite{lu2019monocular, pan2019crossview}, while succeeding approaches introduce knowledge of camera intrinsics to learn the transformation~\cite{roddick2020predicting, philion2020lift,saha2021enabling}. More recent studies adopt transformer-based architectures~\cite{vaswani2017attention}, either using camera geometry~\cite{zhou2022cross, gong2022gitnet, saha2022translating} or not~\cite{bartoccioni2022lara, peng2023bevsegformer}. Somewhat closer to our approach, SkyEye~\cite{gosala2023skyeye} proposes a self-supervised method that exploits spatio-temporal consistency and monocular depth estimation to generate pseudo-labeled BEV maps.
In this paper, we address vision-based semantic BEV segmentation exploiting camera geometry and using a transformer-based architecture. In contrast with previous work though, we do not train our architecture on a specific task, but target zero-shot projection %of new classes 
of \textit{any first-person modality}. 

\myparagraph{Multi-task learning in BEV projection}
Compact BEV representations  
can support multiple downstream tasks, such as object detection, map segmentation and motion planning. A shared backbone network can save computation cost and improve efficiency~\cite{shikun19endtoend}. Thus several works, especially in autonomous driving, use a unified framework to conduct multiple tasks simultaneously~\cite{Yogamani_2019_ICCV, philion2020lift, hu2021fiery}.
The multi-task approach proposed in $M^2$BEV~\cite{vie22m2bev} is based on a BEV representation that makes the uniform depth assumption and thus simplifies the projection process. Some transformer-based methods, such as STSU~\cite{can21structured} and PETRv2~\cite{liu23petrv2}, introduce task-specific queries that interact with shared image features for different perception tasks. BEVFormer~\cite{li22bevformer} first projects multi-view images onto the BEV plane through dense BEV queries and then adopts different task-specific heads over the shared BEV feature map. 
Our models that use auxiliary losses can be considered to fall under the multi-task learning paradigm, where one task is to learn to segment navigable space and/or obstacles in the BEV plane, while the other is to learn to project any feature in BEV.

%\input{sota_old}

%%%%%%%%%%%%%%%%%%%%%%%%%%%%%%%%%%%%%%%%%%%%%%%%%%%%%%%%%%%%%%%%%%
%%%%%%%%%%%%%%%%%%%%%%%%%%%%%%%%%%%%%%%%%%%%%%%%%%%%%%%%%%%%%%%%%%
\section{Disentangling geometry and modality} 
\label{sec:FP_BEV}

\myparagraph{BEV estimation}
the objective of the classical BEV estimation task is to take monocular visual observations 
$\mathbf{I}\rgb\in\mathbb{R}^{W{\times}H{\times}3}$ of size $W{\times}H$, and learn a mapping $\phi$ to translate them into BEVs of varying modality, $\mathbf{M}{=} \phi(\mathbf{I}), \ \mathbf{M}\in\mathbb{R}^{W'{\times}H'{\times}K}$, where $K$ is the dimension of the modality for a single cell of the map. This problem consists of two parts: 
(i) understanding scene semantics e.g. detecting occupancy from color input, and (ii) solving the geometric problem, which requires assigning pixel locations in the FPV to cell positions in the BEV map. In its simplest form, the latter corresponds to an inverse perspective projection, which can be solved in a purely geometric way when cameras are calibrated and depth is available. However, similar to a large body of recent work, eg.~\cite{saha2022translating}, we suppose that depth is \textit{not} available for this mapping. This has essentially two reasons: first, in many situations depth sensors are not applicable or not reliable. Secondly, generalizing the underlying correspondence to forms beyond inverse perspective projections allows to learn more complex visual reasoning processes, for instance to exploit spatial regularities in scenes to predict content occluded or unseen in the FPV, eg. navigable spaces behind objects etc. --- this will be shown in the experimental section. Beyond the inference of unseen scene elements, spatial regularities also play a role in solving the simpler and more basic inverse perspective projection problem itself, which is ill-posed in the absence of depth information~\cite{szeliski2011computer}. 

\begin{figure}[t]  \centering
    \includegraphics[width=0.9\linewidth]{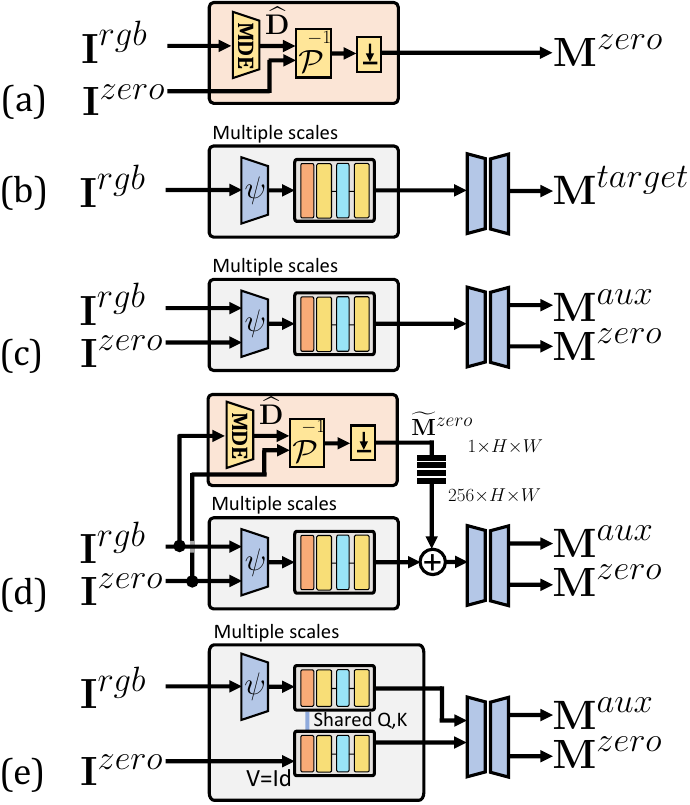}
    \vspace*{-2mm}
    \caption{\label{fig:models}\textbf{Models}: (a) geometric solution based on monocular depth estimation (MDE), inverse projection $\mathcal{P}^{-1}$ and pooling to the ground; (b) end-to-end training to predict a target modality (not zero-shot capable); (c) Zero-BEV model, including optional auxiliary supervision, with feature extractor $\psi$, transformer, and U-Net; (d) Zero-BEV Residual, featuring the geometric solution;  (e) model using inductive bias for disentangling (Section \ref{ssec:inductivebias}).
    }
    \vspace{-1mm}
\end{figure}

\myparagraph{The zero-shot projection task}
consists in taking an image $\mathbf{I}\zero$ from any modality available in FPV beyond RGB, eg. semantic segmentation, optical flow etc., and mapping it to the corresponding BEV map $\mathbf{M}\zero$. This mapping is purely geometric since it does not modify the nature of the content, keeping the modality but changing the view point. The FPV input $\mathbf{I}\zero$ might not contain sufficient information for this projection, i.e. in case where bounding boxes are projected from FPV to BEV, so we suppose the existence of an associated FPV RGB image $\mathbf{I}\rgb$, giving
\setlength{\abovedisplayskip}{5pt}
\setlength{\belowdisplayskip}{5pt}
\begin{equation}
    \mathbf{M}\zero  = \phi\zero (\mathbf{I}\rgb, \mathbf{I}\zero)
\end{equation}
where $\phi\zero$ is the targeted mapping.
The zero-shot nature of the task means that we do \textit{not} require a labelled dataset of pairs $(\mathbf{I}\zero,\mathbf{M}\zero)$ of the targeted modality during training. After training, \textit{any} unseen modality can be projected. % from FPV to BEV. 
A natural geometric baseline uses estimated depth, inverse projection $\mathcal{P}^{-1}$ and pooling to the ground, Figure \ref{fig:models}(a), but this method cannot infer BEV structures invisible in FPV.

We assume access to a dataset of 3D scenes which can be used in simulation, eg. 
Matterport 3D~\cite{chang2018matterport3d} 
or HM3D~\cite{ramakrishnan2021hm3d}. We use these scenes to render procedurally generated pseudo-random data, details will be given in Section \ref{ssec:disentangletrain}. Additionally, and \textit{optionally}, we consider data available for some chosen modality useful for training auxiliary losses, which we call \textit{auxiliary stream}. These optional data maps visual input $\mathbf{I}\aux$ to an output modality $\mathbf{M}\aux$, for which supervision is available during training, occupancy in our experiments.

In the following sections we introduce the functional form of the model, which  will serve as a basis for understanding our main contributions, which are two ways to achieve disentanglement: Section~\ref{ssec:disentangletrain} deals with disentanglement through a specific data generation procedure, whose detailed architecture is given in Section~\ref{ssec:architecture}. Section~\ref{ssec:inductivebias} introduces optional disentanglement through inductive biases.

\subsection{Functional form of the model}
\label{sec:model}

We first introduce the typical form of a single stream model, Figure~\ref{fig:models}(b), which takes an input image $\mathbf{I}\rgb$% (RGB visual input)
and maps it to a BEV map $\mathbf{M}$ of a target modality and omit superscripts when convenient. Adaptations to the zero-shot case will be provided later. 
We consider a backbone network $\psi$, which extracts features in the form of a tensor $\mathbf{H} = \psi(\mathbf{I})$% --- see also Figure~\ref{fig:teaser}
. The full mapping $\phi$ is then given by training a model 
\setlength{\abovedisplayskip}{5pt}
\setlength{\belowdisplayskip}{5pt}
\begin{equation}
\mathbf{M} = \phi(\mathbf{I}) = \phi'(\psi(\mathbf{I})) = \phi'(\mathbf{H}),
\label{eq:nodisentanglement}
\end{equation}
which is supervised with a ground-truth BEV map $\mathbf{M}^*$.
The feature backbone $\psi$ is convolutional and maintains the spatial structure and first-person viewpoint of the input image.

Learning the mapping from FPV to BEV requires to assign a position in polar coordinates $(\theta, \rho)$ in the BEV image for each Cartesian position $(x,y)$ in the first-person tensor $\mathbf{H}$. As in~\cite{roddick2020predicting, philion2020lift,saha2022translating}, we use the intrinsics of a calibrated camera to solve the correspondence between image column $i$ and polar ray $\theta$, and we do not use depth to solve for the rest, but solve this ambiguity via learning.
In the lines of~\cite{saha2022translating}, the heavy lifting of the FPV to BEV correspondence calculations is performed by a sequence of cross-attention layers, on which we focus the formalization below. Additional parts of the model, not directly related to the contributions, will be provided in Section~\ref{ssec:architecture} and the supplementary material.

We decompose the tensor $\mathbf{H}$ into columns and the BEV image $\mathbf{M}$ into polar rays, and solve the assignment problem \textit{pixel height y} $\leftrightarrow$ \textit{cell radius $\rho$ on ray} for each pair (\textit{column}, \textit{ray}) individually. In what follows we drop indices over columns and rays and consider the mapping of a column $\mathbf{h}\in\mathbf{H}$ to a ray $\mathbf{m}\in\mathbf{M}$. Each element $\mathbf{h}_y$ of column $\mathbf{h}$ is a feature vector corresponding to a position $y$ in the column, and each $\mathbf{m}_\rho$ is the map element on position $\rho$ of the ray, whose positional encoding we denote as $\mathbf{p}_{\rho}$.

The assignment problem is solved through Query-Key-Value cross-attention~\cite{vaswani2017attention}, where positional encodings $\mathbf{p}_{\rho}$ on the ray query into the possible positions $\mathbf{h}_y$ on the column they can attend to 
with projections
$Q=\mathbf{p}_{\rho}^T \mathbf{W}_{Q}, 
K=\mathbf{h}_{y}^T \mathbf{W}_{K},$
where $\mathbf{W}_Q$ and $\mathbf{W}_K$ are trainable weight matrices.
For each attention head, this leads to an attention distribution $\alpha_{\rho} = \{ \alpha_{y,\rho} \} $ for each query $\rho$ over attended column positions $y$, calculated classically as in transformer models,
\setlength{\abovedisplayskip}{5pt}
\setlength{\belowdisplayskip}{5pt}
\begin{equation}
\alpha_{y, \rho}=\frac{\textrm{exp}~ {e_{y, \rho}}}{\sum_k \textrm{exp}~ {e_{k, \rho}}},
\
e_{y, \rho}=\frac{
\left(
\mathbf{p}_{\rho}^T \mathbf{W}_{Q}
\right)^T 
\left(\mathbf{h}_{y}^T \mathbf{W}_{K}
\right)
}{\sqrt{D}}.
\label{eq:attention}
\end{equation}
The resulting cell content $\mathbf{m}_\rho$ of the BEV map is then the value projection weighted by attention,
\setlength{\abovedisplayskip}{5pt}
\setlength{\belowdisplayskip}{5pt}
\begin{equation}
\textstyle{
\mathbf{m}_\rho=
\sum_y \alpha_{y,\rho }\mathbf{h}_y^T W_V
}
\label{eq:attentionoutput}
\end{equation}
where we ignore for the moment additional components of a typical transformer architecture, such as multi-head attention, feed-forward and normalization layers.

\begin{figure}[t]  \centering
    \includegraphics[width=0.9\linewidth]{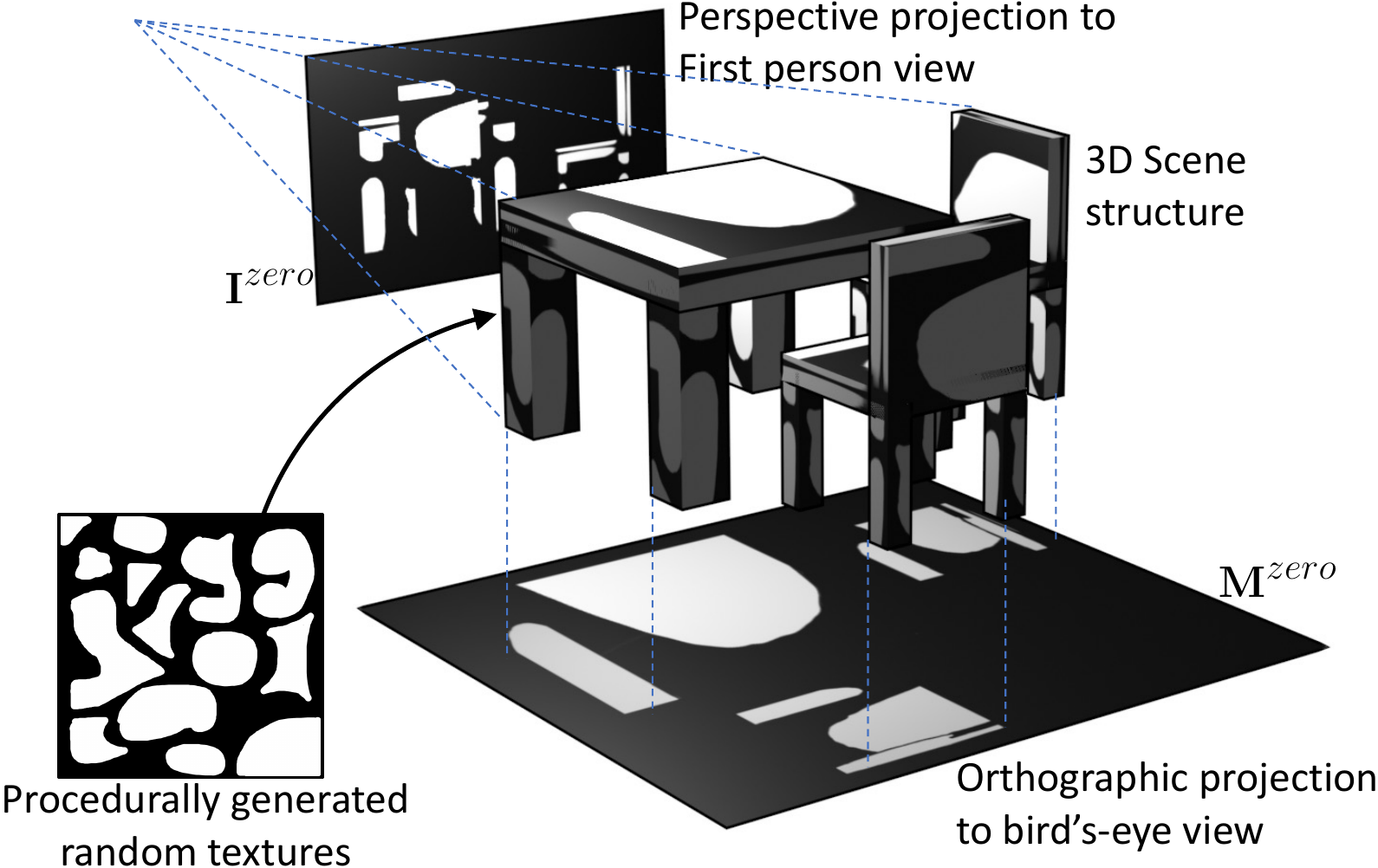}
    \vspace*{-2mm}
    \caption{\label{fig:table}\textbf{Data generation:} we project procedurally generated random textures onto the 3D scene structure and then render the textured mesh into image pairs (first-person view, bird's-eye view) with perspective and orthographic projection, respectively.}
\end{figure}

\subsection{Training for disentanglement}
\label{ssec:disentangletrain}

We introduce a new training procedure, which achieves disentanglement through a dedicated loss combined with synthetically generated data with targeted statistical properties. We combine the main visual input, FPV image $\mathbf{I}\rgb$, with a synthetic pseudo-random \textit{zero-shot data stream} of pairs, FPV input $\mathbf{I}\zero$ and BEV output $\mathbf{M}\zero$, leading to data triplets $(\mathbf{I}\rgb,\mathbf{I}\zero,\mathbf{M}\zero)$ with the following three key properties:
\begin{description}[labelindent=0mm,leftmargin=6mm,topsep=1mm,parsep=0mm,labelsep=0.2em]
    \item[P1:] The two streams are geometrically aligned, i.e. for each sample, $\mathbf{I}\zero$ is taken from the same viewpoint as $\mathbf{I}\rgb$.
    \item[P2:] The output BEV maps $\mathbf{M}\zero$ are labelled synthetically with geometric transforms using the ground-truth 3D scene structure. This structure and overlaid textures are used to generate input-output pairs through perspective and orthographic projections to obtain the FPV image $\mathbf{I}\zero$ and BEV map $\mathbf{M}\zero$ respectively, as in Figure \ref{fig:table}.
    \item[P3:] The content of the zero-shot stream is decorrelated from the content of the primary stream \textit{up to the 3D scene structure}. To be more precise, the texture on the 3D scene structure is binary, pseudo-random, procedurally generated, and does not depend on the scene properties.
\end{description}
Data triplets $(\mathbf{I}\rgb,\mathbf{I}\zero,\mathbf{M}\zero)$ are used for training the mapping function $\phi$, shown in Figure~\ref{fig:models}(c):
\setlength{\abovedisplayskip}{5pt}
\setlength{\belowdisplayskip}{5pt}
\begin{equation}
    \hat{\mathbf{M}}\zero = 
    \phi 
    \left ( \left [ \mathbf{I}\rgb, \mathbf{I}\zero \right ] \right ) = \phi' 
    \left ( \psi \left ( \left [ \mathbf{I}\rgb, \mathbf{I}\zero \right ] \right )\right ),
    \label{eq:singlestream}
\end{equation}
where $[\cdot,\cdot]$ indicates image channel concatenation. 
The network $\phi$ is trained with Dice loss~\cite{milletari2016v}, 
$\mathcal{L}_{D}  ( \hat{\mathbf{M}}\zero, \mathbf{M}\zero )$.
When testing in zero-shot setting, the input pair $(\mathbf{I}\rgb,\mathbf{I}\zero)$ is mapped to the output $\hat{\mathbf{M}}\zero$ through Eq.~(\ref{eq:singlestream}).

\begin{figure}[t]  \centering
    \includegraphics[width=0.9\linewidth]{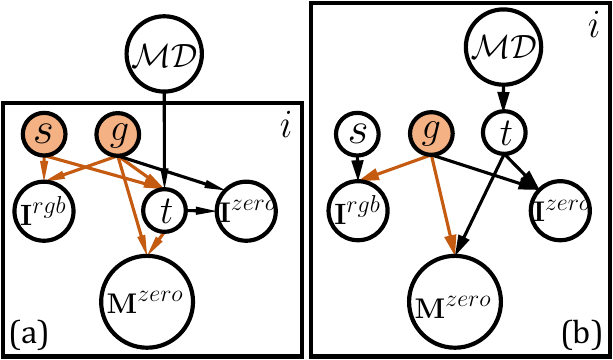}
    \vspace*{-2mm}
    \caption{\label{fig:causal}\textbf{Causal properties} of the data generation process: for each sample $i$, $\mathbf{I}\rgb$ depends on scene geometry $g$ and semantics $s$. \textcolor{orange}{Confounders} between $\mathbf{I}\rgb$ and $\mathbf{M}\zero$ are shaded in orange. $\mathcal{MD}$ is a modality definition, resulting in texture $t$. (a) A fixed meaningful $\mathcal{MD}$ leads to undesired confounders between $\mathbf{I}\rgb$ and  $\mathbf{M}\zero$. (b) We vary $\mathcal{MD}$ over samples and keep it independent of the scene properties $(s,g)$, so the only confounder is geometry $g$.
    }
\end{figure}

\myparagraph{Statistical decorrelation is crucial} property \textbf{P3} is a key design choice for learning correct disentanglement. At first one might be tempted to choose an available modality as training zero-shot data instead of pseudo-random data, for instance semantic segmentation available on the HM3DSem dataset~\cite{yadav2023habitat}. However, this would introduce regularities in the input pair $(\mathbf{I}\rgb, \mathbf{I}\zero)$ due to correlations caused by the confounding scene structure and semantics. This can be formalized in a causal model of the data generation process, shown in Figure \ref{fig:causal}. For each training sample $i$, the input image $\mathbf{I}\rgb$ depends on scene geometry $g$ (including viewpoints) and on scene semantics $s$ (including material properties etc.). \textcolor{orange}{Confounders} between $\mathbf{I}\rgb$ and $\mathbf{M}\zero$ are shaded in orange. The BEV output $\mathbf{M}\zero$ depends on the modality definition $\mathcal{MD}$, which influences the ``texture'' $t$ on the 3D geometry projected to the ground. (a) If during training $\mathcal{MD}$ is fixed to a semantically meaningful modality, eg. occupancy, semantic segmentation etc., then $t$ depends on the scene properties $(s,g)$, leading to unwanted confounders between input $\mathbf{I}\rgb$ and output $\mathbf{M}\zero$, decreasing the role of $\mathbf{I}\zero$. In the worst case scenario, $\mathbf{I}\zero$ becomes unnecessary to predict $\mathbf{M}\zero$, making zero-shot predictions not only degraded but impossible. (b) In our method, the modality definition varies over samples $i$ and is independent of scene properties $(s,g)$. The only confounder between $\mathbf{I}\rgb$ and $\mathbf{M}\zero$ is scene geometry $g$, therefore texture $t$ is necessary for the prediction of $\mathbf{M}\zero$. This leads to correctly taking into account $\mathbf{I}\zero$ to learn the mapping $\phi$, avoiding the exploitation of undesired shortcuts.

\myparagraph{Data generation} 
Training data is generated by applying pseudo-random binary 2D texture images to scene meshes. To make the data semi-structured, avoiding salt and pepper noise, we compose textures from the DTD Dataset \cite{cimpoi14describing}, threshold them with random values and apply random dilation and erosion operators. The resulting structures are dilated until a certain proportion of white pixels is obtained. These textures are rendered over the scene 3D structure and captured with pinhole and orthographic sensors to generate the FPV and BEV map respectively, as shown in Figure~\ref{fig:table}. This data generation process is done \textit{only once} at training, and its details are given in the supplementary material. The proportion of white pixels in the binary images is an important hyper-parameter, and we have explored varying it. A sensitivity analysis is given in the experimental section.

\subsection{Architecture and auxiliary losses}
\label{ssec:architecture}
Our base architecture is inspired by \cite{saha2022translating} and shown in Figure~\ref{fig:models}(c): channel-concatenated input images $[\mathbf{I}\rgb,\mathbf{I}\zero]$ are fed to the convolutional backbone $\psi$, followed by the transformer model described in Section~\ref{sec:model}, and a U-Net which further processes the BEV map and can model inter-ray regularities that the ray-wise transformer cannot deal with. Exact architectures are given in the supplementary material.

\myparagraph{Auxiliary supervision} we augment the base model with an auxiliary supervision of additional modality $\mathbf{M}\aux$ available during training, in our case a binary occupancy map, computed from privileged information in simulation. 
Equation (\ref{eq:singlestream}) extends to $[\hat{\mathbf{M}}\zero, \hat{\mathbf{M}}\aux ] = \phi([ \mathbf{I}\rgb, \mathbf{I}\zero ])$, with mapping $\phi$ trained with Dice loss for both predictions.

\myparagraph{Residual variant}shown in Figure \ref{fig:models}(d), adds a zero-shot component based on monocular depth estimation (MDE) followed by inverse perspective projection $\mathcal{P}^{-1}$ of the zero-shot modality and pooling to the ground, 
$
    \widetilde{\mathbf{M}}\zero = \max_y [ \mathcal{P}^{-1}(\textit{MDE}(\mathbf{I}\rgb), \mathbf{I}\zero) ], 
$
where $\max_y$ indicates max pooling over the vertical dimension. The predicted BEV map is passed through an embedding layer and the resulting 256-channel tensor is added to the transformer output and fed to the U-Net. This residual variant combines the power of state-of-the-art MDE models (we use~\cite{eftekhar2021omnidata}) and the advantages of an end-to-end trained model, which can infer unobserved and occluded information, which is beyond the capabilities of methods based on inverse projection.

\subsection{Disentanglement through inductive biases}
\label{ssec:inductivebias}

We explore the introduction of an inductive bias into the mapping $\phi$ favoring disentangling.
We first study a proof of concept and propose a network based on cross-attention alone, i.e. without initial feature extraction $\psi$ 
and without additional U-Net. It takes the input-output quadruplet including the auxiliary supervision introduced in Section \ref{ssec:architecture} and organizes it into two streams, a zero-shot stream predicting $\mathbf{M}\zero$ from $\mathbf{I}\zero$ and an aux stream predicting $\mathbf{M}\aux$ from $\mathbf{I}\rgb$. 
Both streams share the same \textit{attention distribution} but not the full \textit{transformer layer}. The two streams have different value projections, in particular the \textit{Value} projection of the zero-shot stream is set to the identity function:
\setlength{\abovedisplayskip}{3 pt}
\setlength{\belowdisplayskip}{3pt}
\setlength{\arraycolsep}{0pt}
\begin{eqnarray}
% \begin{array}{lll}
    \hat{\mathbf{M}}\aux = & \phi_{CA} (Q{=}\mathbf{p}(\mathbf{M}\zero), &K{=}\mathbf{I}\rgb, V{=}\mathbf{I}\aux) 
    \label{eq:twostream1} \\
    \hat{\mathbf{M}}\zero = & \phi_{CA} (Q{=}\mathbf{p}(\mathbf{M}\zero), &K{=}\mathbf{I}\rgb, V{=}id(\mathbf{I}\zero)), ~~
    \label{eq:twostream2}    
% \end{array}
\end{eqnarray}
where $\phi_{CA}$ is the cross-attention layer introduced in Sec.~\ref{sec:model}, $\mathbf{p(M)}$ are positional encodings, and query, key projections $Q,K$ are shared between the two streams. $Q,K,V$ are trainable projections, except for $V$ of the zero stream, set to identity (= input $\mathbf{I}\zero$). We then give the following result:
\begin{thm}[Disentangling property]
Let a neural network 
with cross-attention as in Eqs.~(\ref{eq:twostream1})-(\ref{eq:twostream2}) and defined column/ raywise as in Eqs.~(\ref{eq:attention})-(\ref{eq:attentionoutput}) be trained on two streams of (FPV, BEV) pairs of different modalities, where pooling over the vertical dimension is done with a linear function (eg. average). Then, any computation responsible for assigning FPV pixel positions to BEV cells, i.e. the geometric correspondence problem, is restricted to the shared and trainable Key and Query projections, i.e. attention computation.
\label{thm:main}
\end{thm}

\begin{table*}[t] \centering
{\small
\setlength{\tabcolsep}{1pt}
\setlength{\aboverulesep}{0pt}
\setlength{\belowrulesep}{0pt}
\resizebox{\textwidth}{!}{
\begin{tabular}{ a a | c c  | c c c c c c c c | c c }
\toprule 
\rowcolor{GrayBorder}
& {\textbf{Method}} &  {\textbf{\# Params.}}&  {\textbf{Train Set Size}} &  {  \textbf{wall}} & {  \textbf{floor}} & {  \textbf{chair}} & {  \textbf{sofa}} & {  \textbf{bed}} & {  \textbf{plant}}& {  \textbf{toilet}} & {  \textbf{TV}} & {  \textbf{Avg.}}& {  \textbf{Pix Avg.}}  \\
\midrule

\midrule
\cellcolor{green!15}\textbf{(a.1)} & \cellcolor{green!15}\textbf{$\mathcal{P}^{-1}$ w. GT depth \textit{(not comparable/Oracle)}}&  {0} & - &  
14.8 & 32.3 & 41.4 & 43.2 & 36.0 & 45.0 & 21.3 & 28.7 & 32.8 & 42.7
\\
\midrule
\cellcolor{yellow!25}\textbf{(b.1)} &\cellcolor{yellow!25}\textbf{VPN~\cite{pan2019crossview} \textit{(not comp./not zero-shot)}} &  {15M}& 290k & 
 3.7  &  61.9  &  15.0  &  26.4  &  35.9  &  7.0  &  13.8  &  6.6  &  21.3  &  61.1 
\\
\cellcolor{yellow!25}\textbf{(b.2)} &\cellcolor{yellow!25}\textbf{TIM~\cite{saha2022translating} \textit{(not comp./not zero-shot)}} &  {41M}& 290k & 
 5.8  &  67.5  &  21.3  &  30.5  &  39.5  &  9.8  &  16.2  &  7.0  &  24.7  &  64.9 
\\
\cellcolor{yellow!25}\textbf{(b.3)} &\cellcolor{yellow!25}\textbf{TIM~\cite{saha2022translating} w. Sem.Seg. \textit{(not comp./not zero-shot)}} &  {41M}& 290k & 
 7.4  &  69.7  &  30.2  &  42.3  &  46.3  &  15.5  &  20.6  &  11.0  &  30.4  &  68.3 
\\
\midrule
\cellcolor{blue!15}\textbf{(a.2)} &\cellcolor{blue!15}\textbf{$\mathcal{P}^{-1}$ w. learned MDE~\cite{eftekhar2021omnidata}} &  {123M} & {12M}+1M &  
 \cellcolor{orange!15}\underline{9.8}  &  32.2  &  \cellcolor{orange!15}\underline{26.0}  &  34.7  &  31.1  &  \cellcolor{ red!20}\textbf{18.1}  &  \cellcolor{ red!20}\textbf{15.2}  &  \cellcolor{ red!20}\textbf{15.6}  &  22.8  &  39.7 
\\
\cellcolor{blue!15}\textbf{(c)} & \cellcolor{blue!15}\textbf{Zero-BEV (+aux)}  &  {41M} & 290k & 
 8.1  &  \cellcolor{ red!20}\textbf{58.4}  &  23.6  &  \cellcolor{orange!15}\underline{35.5}  &  \cellcolor{orange!15}\underline{35.1}  &  11.5  &  13.3  &  7.8  &  \cellcolor{orange!15}\underline{24.2} &  \cellcolor{orange!15}\underline{58.2} \\
\cellcolor{blue!15}\textbf{(d)} &\cellcolor{blue!15}\textbf{Zero-BEV (+aux, residual)}  &  {123M+41M} & {12M}+1M+290k & 
 \cellcolor{ red!20}\textbf{12.1}  &  \cellcolor{orange!15}\underline{58.2}  &  \cellcolor{red!20}\textbf{27.5}  &  \cellcolor{ red!20}\textbf{39.9}  &  \cellcolor{ red!20}\textbf{39.8}  &  \cellcolor{orange!15}\underline{16.9}  &  \cellcolor{orange!15}\underline{14.5}  &  \cellcolor{orange!15}\underline{12.4}  &  \cellcolor{ red!20}\textbf{27.7}  &  \cellcolor{ red!20}\textbf{58.9} \\ %GTD train, MDE Finetune, MDE Test
\bottomrule
\end{tabular}
}}
\vspace*{-2mm}
\caption{\label{tab:results_main}\textbf{Zero-shot performance of different methods}, reporting IoU on test semantic BEV images unseen during training. The letters correspond to the models in Fig.~\ref{fig:models}. The best values for \tcbox[on line,colframe=white,boxsep=0pt,left=1pt,right=1pt,top=1pt,bottom=1pt,colback=blue!15]{zero-shot methods} are in \tcbox[on line,colframe=white,boxsep=0pt,left=1pt,right=1pt,top=1pt,bottom=1pt,colback=red!20]{\textbf{bold}}, second best are \tcbox[on line,colframe=white,boxsep=0pt,left=1pt,right=1pt,top=1pt,bottom=0pt,colback=orange!10]{\underline{underlined}} --- other approaches are not comparable because they \tcbox[on line,colframe=white,boxsep=0pt,left=1pt,right=1pt,top=1pt,bottom=1pt,colback=green!15]{use Oracle data} or \tcbox[on line,colframe=white,boxsep=0pt,left=1pt,right=1pt,top=1pt,bottom=1pt,colback=yellow!25]{are not zero-shot capable}. Methods (a.2) and (d) use MDE, i.e. additional 12M image tuples for training \cite{eftekhar2021omnidata} + 1M RGB-depth image pairs for finetuning.
}
\end{table*}

\begin{table*}[t] \centering
{\footnotesize
\setlength{\tabcolsep}{1pt}
\setlength{\aboverulesep}{0pt}
\setlength{\belowrulesep}{0pt}
{
\begin{tabular}{ a a | c c c c c c c c | c c }
\toprule 
\rowcolor{GrayBorder}
& {\textbf{Model}} & {\textbf{wall}}&{  \textbf{floor}} & {  \textbf{chair}} & {  \textbf{sofa}} & {  \textbf{bed}} & {  \textbf{plant}}& {  \textbf{toilet}} & {  \textbf{TV}} & {  \textbf{Avg.}}& {  \textbf{Pix Avg.}}  \\
\midrule

\midrule
\cellcolor{gray!15}\textbf{(-)} & \cellcolor{gray!15}\textbf{Zero-BEV}&   
 7.2  &  57.1  &  21.7  &  33.5  &  33.8  &  10.1  &  12.3  &  6.3  &  22.8 &  57.7 \\
\midrule
\cellcolor{blue!15}\textbf{(c)} & \cellcolor{blue!15}\textbf{Zero-BEV + aux (=obstacles + navigable)} & 
  8.1  &  \cellcolor{orange!15}\underline{58.4}  &  23.6  &  35.5  &  35.1  &  11.5  &  13.3  &  7.8  &  24.2 &  \cellcolor{orange!15}\underline{58.2} \\
\cellcolor{gray!15}\textbf{(c.I)} & \cellcolor{gray!15}\textbf{Zero-BEV + aux (=obstacles)} & 
 8.1  &  58.3  &  23.5  &  35.7  &  \cellcolor{orange!15}\underline{35.3}  &  10.9 &  12.0 &  6.8 &  23.8  &  57.1  \\ 
\cellcolor{gray!15}\textbf{(c.II)} & \cellcolor{gray!15}\textbf{Zero-BEV + aux (=navigable)} & 
 8.0  &  58.0  &  23.0  &  34.9  &  34.9  &  9.9 &  12.2 &  5.9 &  23.4  &  57.8  \\ 
\midrule
\cellcolor{blue!15}\textbf{(d)} & \cellcolor{blue!15}\textbf{Zero-BEV + aux (=obstacles + navigable), residual} &
 \cellcolor{ red!20}\textbf{12.1}  &  58.2  &  \cellcolor{ red!20}\textbf{27.5}  &  \cellcolor{ red!20}\textbf{39.9}  &  \cellcolor{ red!20}\textbf{39.8}  &  \cellcolor{ red!20}\textbf{16.9}  &  \cellcolor{orange!15}\underline{14.5}  &  \cellcolor{orange!15}\underline{12.4}  &  \cellcolor{ red!20}\textbf{27.7}  &  \cellcolor{ red!20}\textbf{58.9} \\ % I propose to repeat line (g) of table 1 here.
\cellcolor{gray!15}\textbf{(d.I)} & \cellcolor{gray!15}\textbf{Zero-BEV, residual} & 
     \cellcolor{orange!15}\underline{10.9}  &  51.2  &  \cellcolor{orange!15}\underline{23.9}  &  \cellcolor{orange!15}\underline{36.2}  &  34.8  &  \cellcolor{orange!15}\underline{16.8}  &  \cellcolor{ red!20}\textbf{14.7}  &  \cellcolor{ red!20}\textbf{14.4}  &  \cellcolor{orange!15}\underline{25.4} &  53.9  \\ 
\midrule
\cellcolor{gray!15}\textbf{(e)} & \cellcolor{gray!15}\textbf{Zero-BEV + aux (=obstacles + navigable), inductive bias} & 
 8.0  &  \cellcolor{ red!20}\textbf{59.0}  &  21.9  &  33.9  &  34.7  &  9.9  &  12.3  &  5.8  &  23.2 &  56.9  \\ 
\bottomrule
\end{tabular}
}}
\vspace*{-2mm}
\caption{\label{tab:models}\textbf{Model variants}: we explore the impact of the auxiliary loss supervising a binary channel defined in different ways, and the impact of the inductive bias. \tcbox[on line,colframe=white,boxsep=0pt,left=1pt,right=1pt,top=1pt,bottom=1pt,colback=blue!15]{Highlighted rows} in this table and the following ones are copied from Table \ref{tab:results_main}.
\vspace*{-2mm}
}
\end{table*}

\noindent
A constructive proof is given in the supplementary material.

This result shows that the inductive bias leads to alignment for disentangling: the attention distribution corresponds to the desired geometric transformation, used by the second stream to project the zero-shot modality to BEV without changing it, since its \textit{Value} projection is set to identity.
While the result holds for pure cross-attention layers, it does not hold in working variants of transformer networks, which include feed-forward layers and other modules. Our model also includes a backbone $\psi$ and a U-Net, shown in Figure \ref{fig:models}(e). For these reasons, this inductive bias alone is not sufficient for disentanglement and we explore a combination with the data generation procedure given in Section \ref{ssec:disentangletrain}.

\section{Experiments}
\label{sec:exp}

\myparagraph{Task and data generation}
Data is generated using the Habitat simulator~\cite{Savva_2019_ICCV} with 
the HM3DSem~\cite{yadav2023habitat} dataset 
consisting of $145$ train scenes and $36$ val scenes. The $4$ minival scenes are used for validation and the remaining $32$ for testing. We target the BEV semantic segmentation task using the 6 semantic categories used in~\cite{chaplot2020object}, \textit{`chair', `sofa', `bed', `potted plant', `toilet', `tv'}, plus \textit{`wall'} and \textit{`floor'}, for a total of 8 categories.
For each scene we collect $2000$ tuples from two points of view: FPV images 
captured with a pinhole camera with resolution $384{\times}384$ and field of view (FOV) $79^{\circ}$, and top-down BEV maps of size $100{\times}100$ captured with an orthographic sensor over the FPV position and facing down, covering an area of $5\textrm{m}{\times}5\textrm{m}$ in front of the FPV sensor. For each sensor we record RGB, depth and semantic annotations. Top-down orthographic depth images are thresholded at a fixed value to generate navigable and obstacle BEV maps used as optional auxiliary modalities. All BEV maps are masked with the FPV FOV, estimated by projecting the FPV depth to the ground and computing its convex hull.

\myparagraph{Implementation and training details}
The feature extractor $\psi$ is a pretrained ResNet50 \cite{he2016deep} followed by 
a feature pyramid~\cite{lin2017feature} to extract FPV feature maps at four resolutions. As in~\cite{saha2022translating}, each FPV feature map is projected to a BEV band independently using the transformer-based network $\phi$ and concatenated to form a BEV feature map of size $256{\times}100{\times}100$. This is processed by a small U-Net akin to \cite{philion2020lift, bartoccioni2022lara} to generate $\mathbf{\hat{M}}\zero$ and optionally $\mathbf{\hat{M}}\aux$. The model is trained with Dice loss~\cite{milletari2016v} and Adam optimizer~\cite{kingma14adam}, batch size $16$ and initial learning rate $lr=1e-4$ with $0.9$ exponential decay and halving of $lr$ on plateau of val performance. Results for this method are in Table~\ref{tab:results_main}(c). We compare with the following baselines and variants:

\begin{description}[noitemsep,labelindent=0mm,leftmargin=4mm,topsep=0mm,labelsep=1mm]
\item[Inverse projection and pooling to ground using depth] of FPV semantic segmentation is used by many existing methods~\cite{chaplot2020object, ramakrishnan2022poni}. We decline it in two flavours: (i) using ground-truth depth from Habitat simulator --- Table~\ref{tab:results_main}(a.1); (ii) Omnidata normalized MDE model~\cite{eftekhar2021omnidata, kar20223d}, finetuned on a custom dataset of $1$M RGB-depth image pairs from HM3D~\cite{ramakrishnan2021hm3d} for metric MDE --- Table~\ref{tab:results_main}(a.2).
\item[View Parsing Network (\textit{VPN})~\cite{pan2019crossview}], a non zero-shot method that does not use camera intrinsics. It is trained to predict target 8-class BEV semantic maps --- Table~\ref{tab:results_main}(b.1). 
\item[Translating Images into Maps (\textit{TIM})~\cite{saha2022translating}], a non zero-shot learning method with architecture similar to ours, Figure~\ref{fig:models}(b), and trained to predict the target BEV semantic maps --- Table~\ref{tab:results_main}(b.2). We also train a model with an additional input channel, the FPV semantic segmentation, so that it has the exact same information of zero-shot methods. This should represent an upper bound when it can be trained on the task, but cannot be used for zero-shot projections of any modality. Results are in Table~\ref{tab:results_main}(b.3).
\item[Zero-BEV residual model] combines geometry with learning, Figure~\ref{fig:models}(d), concatenating BEV feature maps generated by our approach and by projecting with the metric MDE model described above. The resulting features are processed by the same U-Net to zero-shot generate BEV maps. Results are displayed in Table~\ref{tab:results_main}, row (d).
\end{description}

%\noindent
\myparagraph{Metrics} we report IoU in two variants: averaging over classes, and accumulating over pixels (see supp.mat.). The former weights classes equally, whereas the latter gives weights proportional to their appearance in the data.

\begin{figure*}[t]  \centering
    \includegraphics[width=0.99\linewidth]{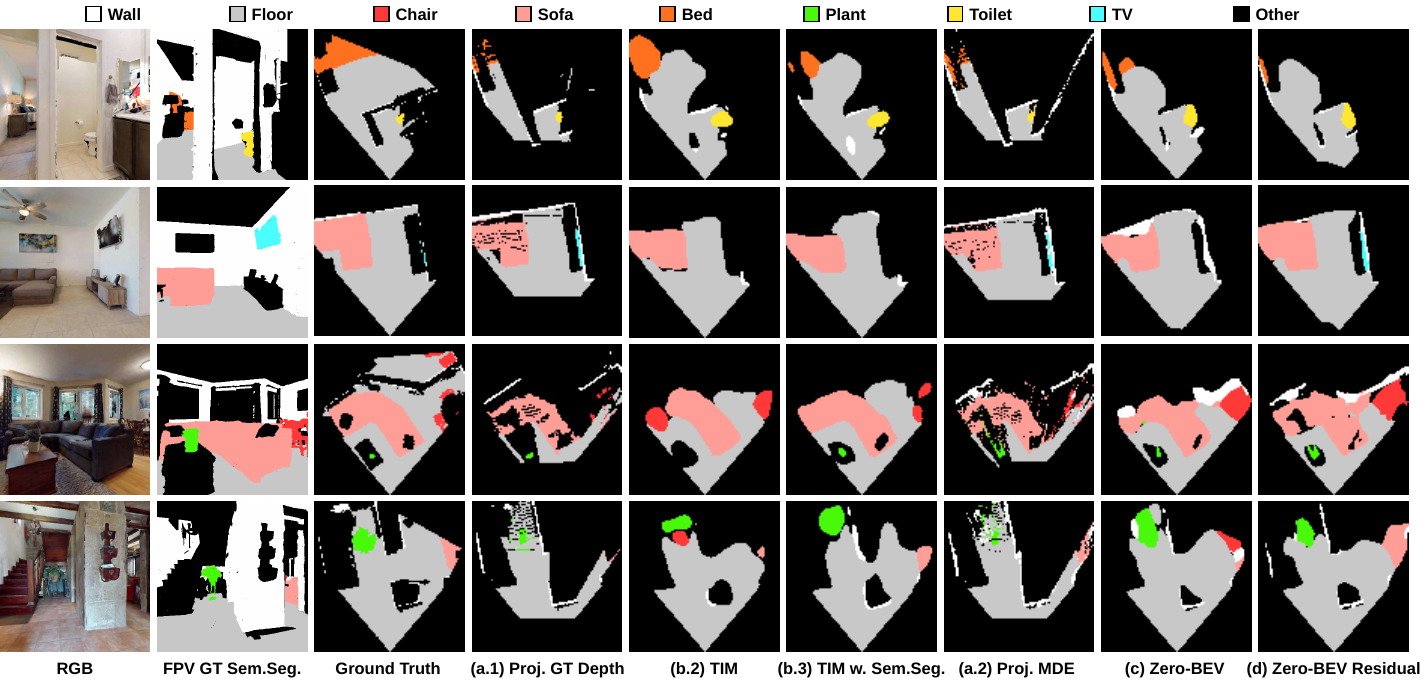}
    \vspace*{-3mm}
    \caption{\label{fig:qualitative}\textbf{Qualitative results} on HM3DSem test scenes. (a.1) uses ground-truth depth and methods (c.2) and (c.3) are not zero-shot capable, thus not comparable. (a.2), (c) and (d) are zero-shot models --- see Table~\ref{tab:results_main}. Zero-BEV models produce significantly better BEV maps.
    %, comparable with the fully supervised models.
    }
    \vspace*{-1mm}
\end{figure*}

\myparagraph{Comparisons with baselines and SOTA} are given in Table \ref{tab:results_main} for projections of semantic segmentations taken from HM3DSem~\cite{yadav2023habitat}. \method{} (c) outperforms the geometric zero-shot capable baseline (a.2), which resorts to inverse perspective projection with learned depth, on both metrics, class IoU and pix IoU and using both the pure and the residual variants. The biggest gains occur for the \textit{floor} class, which can be explained by the power of the end-to-end trained model to infer unseen and occluded floor space, but big gains are also obtained for \textit{wall, chair, sofa} and \textit{bed}. Interestingly, on the pixel IoU metric \method{} even outperform the projection with ground-truth depth, and examples in Figure \ref{fig:qualitative} make it clear why: projecting only observed information from FPV to BEV leaves a large amount of holes in the map. The end-to-end trained methods, Table~\ref{tab:results_main}(b.*), are not zero-shot capable as they have been trained for the target modality for which we evaluate --- the numbers are given \textit{for information}. Compared to these well-received methods, \method{} adds zero-shot capabilities and thus new applications. Interestingly, \method{} outperforms VPN on class IoU and is not far in pixel IoU. The residual variant (d) boosts the performance of \method{} mostly on smaller classes and even outperforms a state-of-the-art \textit{non zero-shot} method like TIM. Figure \ref{fig:qualitative} visually shows how this model obtains hints of the presence of smaller structures from the geometric mapping and is capable of expanding on them.

\myparagraph{Impact of auxiliary losses} is given in Table \ref{tab:models}(c.*). Compared to the base variant (-), the auxiliary loss mainly adds reconstruction quality for smaller semantic classes and does not seem to be responsible for the power to infer unseen information. We explore the usage of different modalities as choice of the binary auxiliary signal and report slight differences. For the residual model (d.*), auxiliary losses also boost mean performance and impact most classes.

\myparagraph{Impact of the inductive bias} is explored in Table~\ref{tab:models}(e). The model is outperformed by the comparable variant (c), which we conjecture is due to two properties: (i) our data is generated by max-pooling vertically to the ground, which is not covered by the disentangling property (Theorem 3.1) and further detailed in the supplementary material; (ii) restricting the cross-attention computations to geometry may help disentangling, but can potentially hurt the expressive power of the network, removing its capacity to model spatial regularities unrelated to the geometric mapping.

\myparagraph{Training the residual model} requires depth, which \textit{at training} can be provided in different ways, investigated in Table~\ref{tab:re_models}. The network in first row is trained using ground-truth from the simulator, the second with depth estimated with the MDE model, and the third is trained first with GT and then MDE depth. All models are tested with MDE predicted depth. The third strategy achieves the best results.

\myparagraph{Qualitative examples} are given in Figure \ref{fig:qualitative}. Projections using depth, either ground-truth (a.1) or estimated with MDE (a.2) are precise on walls and fine structures, but suffer from typical artefacts such as holes and depth quantization noise. Fully supervised TIM~\cite{saha2022translating}, (b.2) and (b.3), can infer regularities in top-down maps. As expected, FPV semantic segmentation, available to (b.3), enhances BEV map quality. Zero-BEV (c) appears qualitatively close to its non zero-shot counterparts. Zero-BEV Residual (d) even seems to improve over TIM variants, exploiting geometric projection for fine elements and learning to regularize BEV structures, and going as far as correcting small inaccuracies in ground-truth labels (eg. the left-most wall on the bottom example).

\begin{table}[t] \centering
{\small
\setlength{\tabcolsep}{1pt}
\setlength{\aboverulesep}{0pt}
\setlength{\belowrulesep}{0pt}
%\resizebox{\textwidth}{!}
{
\begin{tabular}{a a | c c c c c c c c | c c }
\toprule 
\rowcolor{GrayBorder}
 &{\textbf{\scriptsize Train depth}} & {\textbf{\scriptsize wall}}&{  \textbf{\scriptsize floor}} & {  \textbf{\scriptsize chair}} & {  \textbf{\scriptsize sofa}} & {  \textbf{\scriptsize bed}} & {  \textbf{\scriptsize plant}}& {  \textbf{\scriptsize toilet}} & {  \textbf{\scriptsize TV}} & {  \textbf{\scriptsize Avg.}}& {  \textbf{\scriptsize  Pix}}  \\
\midrule
\cellcolor{gray!15}&\cellcolor{gray!15}\textbf{\scriptsize GT} & 
 11.2  &  \cellcolor{orange!15}\underline{57.4}  &  26.0  &  \cellcolor{orange!15}\underline{39.2}  &  \cellcolor{orange!15}\underline{38.5}  &  \cellcolor{orange!15}\underline{16.9}  &  \cellcolor{ red!20}\textbf{15.5}  &  \cellcolor{orange!15}\underline{15.2}  &  \cellcolor{orange!15}\underline{27.5}  &  56.8 \\ %GTD train
\cellcolor{gray!15}&\cellcolor{gray!15}\textbf{\scriptsize Pred.} & 
 \cellcolor{orange!15}\underline{11.7} &  52.0 &  \cellcolor{orange!15}\underline{27.1} &  38.8  &  37.7  & \cellcolor{ red!20}\textbf{18.6} & \cellcolor{orange!15}\underline{15.0} & \cellcolor{ red!20}\textbf{15.5} &  27.1  &  \cellcolor{orange!15}\underline{57.2} \\  % MDE train
\cellcolor{blue!15}{\textbf{\scriptsize(d)}}&\cellcolor{blue!15}\textbf{\scriptsize GT, pred.}&    
 \cellcolor{ red!20}\textbf{12.1}  &  \cellcolor{ red!20}\textbf{58.2}  &  \cellcolor{ red!20}\textbf{27.5}  &  \cellcolor{ red!20}\textbf{39.9}  &  \cellcolor{ red!20}\textbf{39.8}  &  \cellcolor{orange!15}\underline{16.9}  &  14.5  &  12.4  &  \cellcolor{ red!20}\textbf{27.7}  &  \cellcolor{ red!20}\textbf{58.9} \\ %GTD train, MDE Finetune
\bottomrule
\end{tabular}
}}
\vspace*{-2mm}
\caption{\label{tab:re_models}\textbf{Training the residual model}, which requires depth. First line trains with ground-truth depth, second with predicted, third first GT then predicted depth. Testing \textbf{always} uses predicted depth.
}
\end{table}

\begin{table}[t] \centering
{\small
\setlength{\tabcolsep}{1pt}
\setlength{\aboverulesep}{0pt}
\setlength{\belowrulesep}{0pt}
%\resizebox{\textwidth}{!}
{
\begin{tabular}{ a a | c c c c c c c c | c c }
\toprule 
\rowcolor{GrayBorder}
 &{\textbf{\footnotesize Matter}} & {\textbf{\footnotesize wall}}&{  \textbf{\footnotesize floor}} & {  \textbf{\footnotesize chair}} & {  \textbf{\footnotesize sofa}} & {  \textbf{\footnotesize bed}} & {  \textbf{\footnotesize plant}}& {  \textbf{\footnotesize toilet}} & {  \textbf{\footnotesize TV}} & {  \textbf{\footnotesize Avg.}}& {  \textbf{\footnotesize Pix}}  \\
\midrule
\cellcolor{gray!15}&\cellcolor{gray!15}$\textbf{10\%}$&   
 7.9  &  56.0  &  \cellcolor{ red!20}\textbf{23.7}  &  34.3  &  33.9  &  \cellcolor{orange!15}\underline{11.4}  &  12.7  &  \cellcolor{orange!15}\underline{7.7}  &  23.5 &  58.5 \\
\cellcolor{gray!15}&\cellcolor{gray!15}$\textbf{15\%}$ & 
 \cellcolor{ red!20}\textbf{8.7}  &  56.4  &  \cellcolor{orange!15}\underline{23.6}  &  \cellcolor{ red!20}\textbf{35.6}  &  \cellcolor{orange!15}\underline{35.0}  &  11.2  &  12.9  &  7.4  &  \cellcolor{orange!15}\underline{23.9} &  \cellcolor{orange!15}\underline{58.8}  \\
\cellcolor{blue!15}\textbf{(c)}&\cellcolor{blue!15}$\textbf{20\%}$ & 
  \cellcolor{orange!15}\underline{8.1}  &  \cellcolor{orange!15}\underline{58.4}  &  \cellcolor{orange!15}\underline{23.6}  &  \cellcolor{orange!15}\underline{35.5}  &  \cellcolor{ red!20}\textbf{35.1}  &  \cellcolor{ red!20}\textbf{11.5}  &  \cellcolor{orange!15}\underline{13.3}  &  \cellcolor{ red!20}\textbf{7.8}  &  \cellcolor{ red!20}\textbf{24.2} &  58.2 \\
\cellcolor{gray!15}&\cellcolor{gray!15}$\textbf{25\%}$ & 
 7.3  &  \cellcolor{ red!20}\textbf{59.1}  &  23.4  &  35.1  &  34.8  &  10.6  &  12.6  &  6.7  &  23.7 &  58.5  \\
\cellcolor{gray!15}&\cellcolor{gray!15}$\textbf{30\%}$ & 
 8.0  &  57.4  &  23.3  &  34.7  &  \cellcolor{ red!20}\textbf{35.1}  &  10.4  &  \cellcolor{ red!20}\textbf{13.4}  &  6.1  &  23.6 &  \cellcolor{ red!20}\textbf{60.4}  \\
\bottomrule
\end{tabular}
}}
\vspace*{-2mm}
\caption{\label{tab:data_gen}\textbf{Data density}: we analyze the impact of the amount of ``white matter'' (pixels) in the textures mapped on the 3D scene structure for the training data, cf. Section~\ref{ssec:disentangletrain} and Figure~\ref{fig:table}. 
The amount in the FPV/BEV images $(\mathbf{I}\zero,\mathbf{M}\zero)$ may differ.
}
\end{table}

\myparagraph{Data generation} Table \ref{tab:data_gen} provides a sensitivity analysis of the amount of ``\textit{white matter}'' (=pixels) in textures before mapping them on the 3D scene. Sensitivity is low, 20\% was chosen based on avg. IoU. In Table \ref{tab:data_type} we compare the synthetic data generation with two alternative strategies: one based on modifying the existing HM3DSem textures corresponding to semantic labels (\textit{Mod.Sem.}), with the idea of generating distributions of 2D shapes which are close to the existing scene structure, somewhat violating property \textbf{P3}, and heavily modifying the textures with morphological operations. This leads to degraded performance, providing further evidence for the importance of \textbf{P3}. The second one consists in using ground-truth depth to project random binary shapes from FPV to BEV (\textit{Depth proj.}), which, as expected, behaves similarly to depth projection-based methods --- cf. Table~\ref{tab:results_main}(a.*) --- as it does not allow to learn inferring unseen BEV content from FPV images.

\begin{figure}[t]  \centering
    \includegraphics[width=0.95\linewidth]{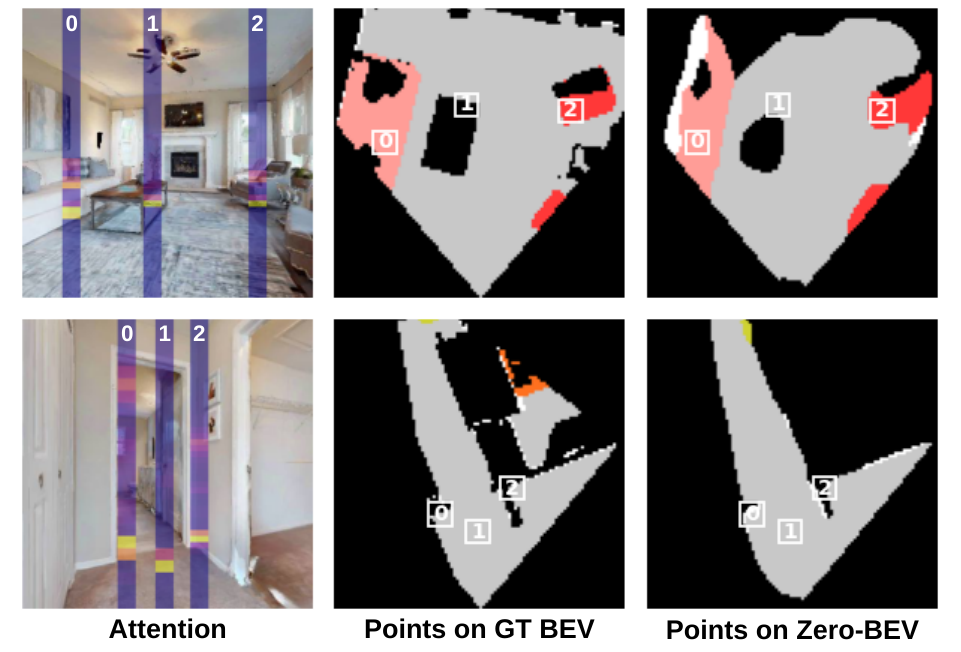}
    \vspace*{-2mm}
    \caption{\label{fig:attention}\textbf{Visualization of attention distributions}: (Right) predicted BEV map with 3 selected positions, numbered; (Middle) GT BEV map; (Left) Corresponding attention distributions for the FPV columns corresponding to the selected map positions.}
    \vspace*{-2mm}
\end{figure}

\myparagraph{Attention distributions} are visualized in Figure \ref{fig:attention} for two example images and three selected map cells for each image. Generally, attention focuses on vertical furniture boundaries, eg. the sofa. Interestingly, the transformer attends to the floor even when the spot is occluded, as is the case for point \#1 in the top image (behind the table) and \#2 in the bottom image (behind the wall). 

\begin{figure}[t]  \centering
    \includegraphics[width=0.92\linewidth]{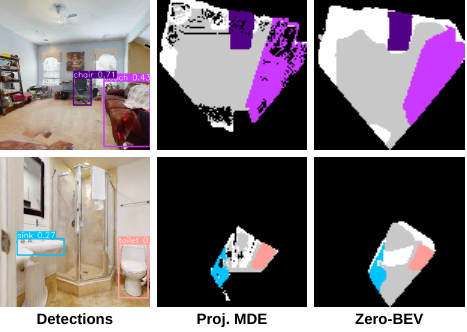}
    \vspace*{-2mm}
    
    \caption{\label{fig:yolo}\textbf{Zero-shot projection of objects} detected in the FPV image (Left); (Middle) geometric projection with learned depth; (Right) projection with our \method{} method.}
    \vspace*{-2mm}
\end{figure}

\begin{figure}[t]  \centering
    \includegraphics[width=0.92\linewidth]{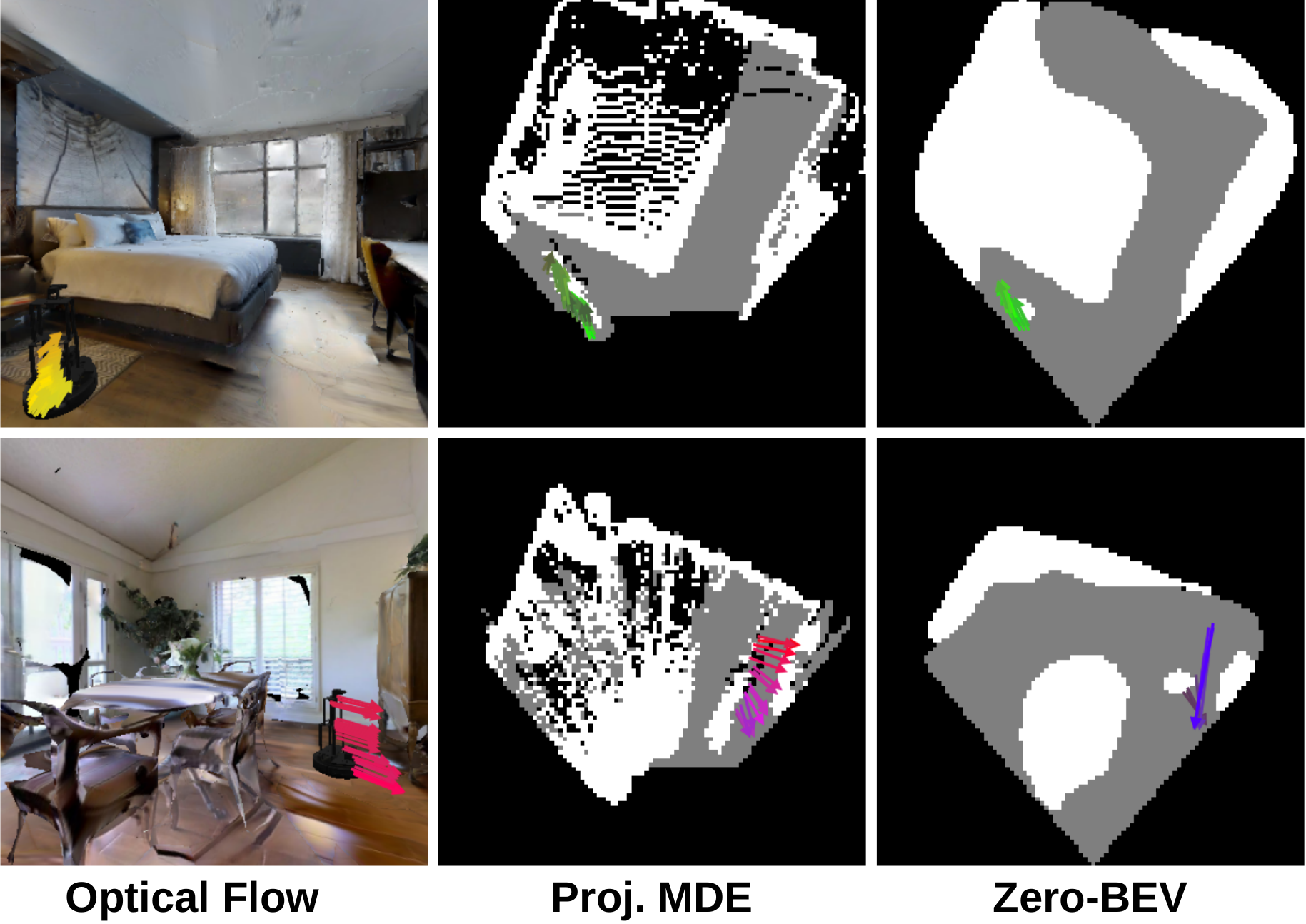}
    \vspace*{-2mm}
    
    \caption{\label{fig:flow}\textbf{Optical Flow} calculated on an FPV pair $(t,t{+}1)$ and displayed on frame $t$ (Left),  then projected to BEV. (Middle) geometric projection with learned MDE; (Right) with Zero-BEV.}
    \vspace*{-2mm}
\end{figure}

\myparagraph{Additional zero-shot capabilities} in Figure \ref{fig:yolo} we zero-shot project \textit{object bounding boxes} detected with YOLO~\cite{YOLOCVPR2016} in FPV. Compared to the geometric baseline, the maps are cleaner and less noisy. In Figure \ref{fig:flow} we zero-shot project \textit{optical flow} detected with RAFT~\cite{teed2020raft}. For each flow vector, two forward passes are performed for binary filled circles put on the starting point and the end point, respectively.  Compared to the geometric baseline, the BEV flow vectors provided by \method{} are more accurate, which we conjecture is due to the difficulty of using geometric projections on finely structured moving objects, here the mobile robot.

\begin{table}[t] \centering
{\small
\setlength{\tabcolsep}{1pt}
\setlength{\aboverulesep}{0pt}
\setlength{\belowrulesep}{0pt}
%\resizebox{\textwidth}{!}
{
\begin{tabular}{ a a | c c c c c c c c | c c }
\toprule 
\rowcolor{GrayBorder}
 &{\textbf{\scriptsize Data}} & {\textbf{\scriptsize wall}}&{  \textbf{\scriptsize floor}} & {  \textbf{\scriptsize chair}} & {  \textbf{\scriptsize sofa}} & {  \textbf{\scriptsize bed}} & {  \textbf{\scriptsize plant}}& {  \textbf{\scriptsize toilet}} & {  \textbf{\scriptsize TV}} & {  \textbf{\scriptsize Avg.}}& {  \textbf{\scriptsize Pix}}  \\
\midrule

\midrule
\cellcolor{blue!15}{\scriptsize\textbf{(c)}}&\cellcolor{blue!15}\textbf{\scriptsize Synth ({\tiny$20\%$})}&   
  \cellcolor{orange!15}\underline{8.1}  &  \cellcolor{orange!15}\underline{58.4}  &  \cellcolor{ red!20}\textbf{23.6}  &  \cellcolor{ red!20}\textbf{35.5}  &  \cellcolor{ red!20}\textbf{35.1}  &  \cellcolor{orange!15}\underline{11.5}  &  \cellcolor{orange!15}\underline{13.3}  &  \cellcolor{ red!20}\textbf{7.8}  &  \cellcolor{ red!20}\textbf{24.2} &  \cellcolor{ red!20}\textbf{58.2} \\
\cellcolor{gray!15}&\cellcolor{gray!15}\textbf{\scriptsize ModSem ({\tiny$30\%$})} & 
 5.0  &  \cellcolor{ red!20}\textbf{60.1}  &  20.4  &  \cellcolor{orange!15}\underline{31.2}  &  \cellcolor{orange!15}\underline{32.3}  &  8.3  &  10.5  &  4.7  &  \cellcolor{orange!15}\underline{21.6} &  \cellcolor{orange!15}\underline{48.9} \\ 
\cellcolor{gray!15}&\cellcolor{gray!15}\textbf{\scriptsize Depth proj.} & 
 \cellcolor{ red!20}\textbf{8.7}  &  32.2  &  \cellcolor{orange!15}\underline{22.5}  &  29.5  &  27.8  &  \cellcolor{ red!20}\textbf{12.4}  &  \cellcolor{ red!20}\textbf{13.9}  &  \cellcolor{orange!15}\underline{7.7}  &  19.3 &  41.7 \\
\bottomrule
\end{tabular}
}}
\vspace*{-2mm}
\caption{\label{tab:data_type}\textbf{Data gen.}: \textit{Synth} is procedurally generated (Section~\ref{ssec:disentangletrain}). \textit{Mod Sem} is based on semantic segmentation modified with morphological operations. The last line projects FPV white rectangles to BEV with ground-truth depth. The amount of white matter (in parentheses) was optimized  on the val set for the first two lines.
}
\end{table}

\section{Conclusion}
\label{sec:conclusion}
Our method for zero-shot projection from FPV to BEV combines the advantages of end-to-end methods to infer unseen information with the zero-shot capabilities of geometric methods based on inverse projection. The key contribution is a new data generation process, which decorrelates 3D scene structure from other scene properties. The method outperforms the competition and is applicable to new tasks.

{
    \small
%    \bibliographystyle{ieeenat_fullname}
%    \bibliography{references,references_chris_zotero}

}
\clearpage
\setcounter{page}{1}
\maketitlesupplementary

\begin{figure*}[t]  \centering
    \includegraphics[width=0.99
    \linewidth]{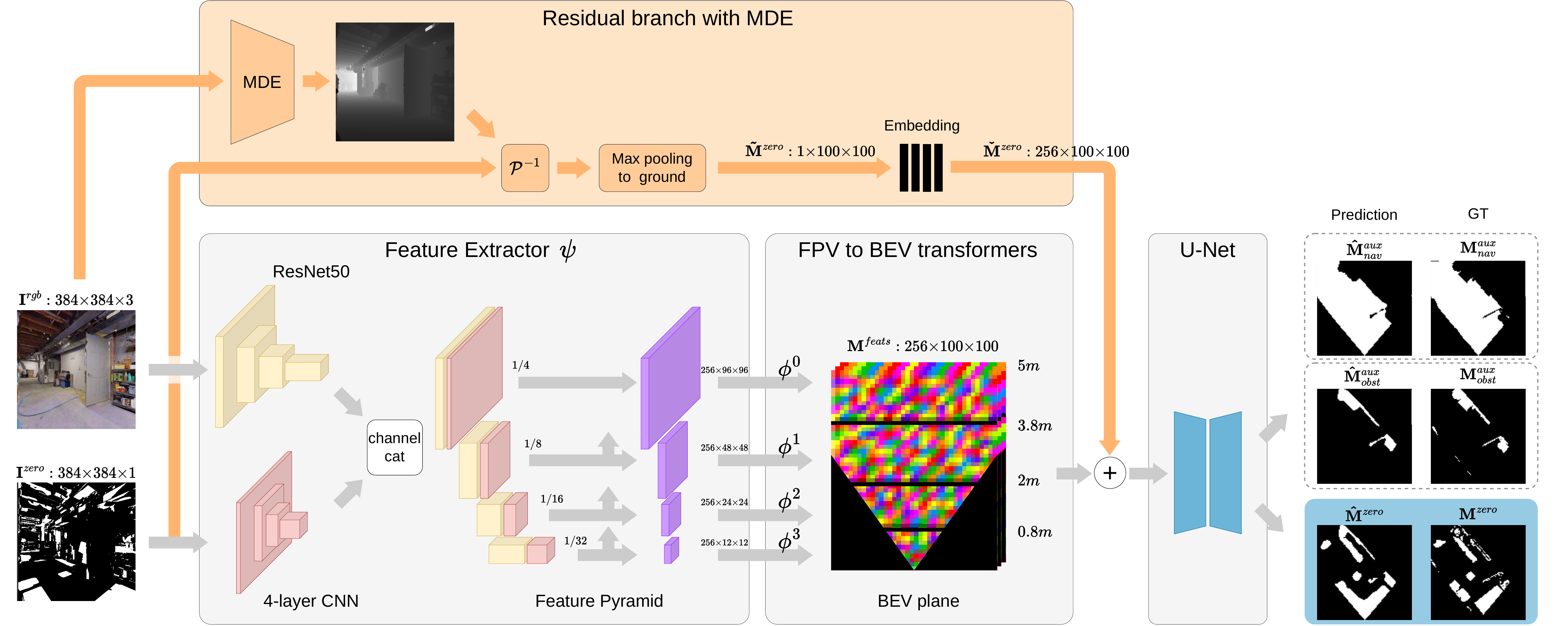}
    \caption{\label{fig:arch}\textbf{Zero-BEV architectures}: the \tcbox[on line,colframe=white,boxsep=0pt,left=1pt,right=1pt,top=1pt,bottom=1pt,colback=gray!15]{base model below in gray}, and the \tcbox[on line,colframe=white,boxsep=0pt,left=1pt,right=1pt,top=1pt,bottom=1pt,colback=orange!15]{residual branch in orange}.
    }
    %\vspace*{-1mm}
\end{figure*}

%\noindent
%\textit{(Unless explicitly stated, all references to Figures, Tables or Bibliography are to material in this supplementary document and not to the main paper)}

%%%%%%%%%%%%%%%%%%%%%%%%%%%%%%%%%%%%%%%%%%%%%%%%%%%%%%%%%%%%%%%%%%%%%%%%%%%%%%%%%%%%%%%%%%%%%%%%%
%%%%%%%%%%%%%%%%%%%%%%%%%%%%%%%%%%%%%%%%%%%%%%%%%%%%%%%%%%%%%%%%%%%%%%%%%%%%%%%%%%%%%%%%%%%%%%%%%
\section{Architecture Details}

\subsection{Zero-BEV base model}

Our base model comprises three main blocks, shown in Figure~\ref{fig:arch}: the feature extractor $\psi$, the first person view (FPV) to bird's-eye view (BEV) transformer-based mapping, and the BEV decoder, a U-Net like model (details further below) which for simplicity we call U-Net.

\myparagraph{Feature extractor $\psi$} converts FPV images $\mathbf{I}$ into multiscale features $\mathbf{H} = \psi(\mathbf{I})$. The RGB image $\mathbf{I}\rgb$ of dimensions $384{\times}384{\times}3$ is processed by a pretrained ResNet50~\cite{he2016deep}, from which we get the output of four intermediate layers, specifically \textsc{layer1}, \textsc{layer2}, \textsc{layer3} and \textsc{layer4}, which produce four feature maps $\mathbf{f}\rgb_b, b=0,\dots,3$ of spatial dimensions $1/4$, $1/8$, $1/16$, $1/32$ of the input image respectively. 

The binary $384{\times}384{\times}1$ zero-shot image $\mathbf{I}\zero$ should ideally not be processed at all, since the goal of the newly proposed zero-shot task is to perform a geometric projection from FPV to BEV and to let the actual modality definition unchanged: do not change the values of the pixels, but only place them on the correct position of the BEV map. In practice, the model needs to subsample $\mathbf{I}\zero$ to match the resolution of the feature maps $\mathbf{f}\rgb_b$ such that these two tensors can be stacked in the subsequent layers. Instead of simply subsampling and pooling, we feed $\mathbf{I}\zero$ to a small CNN with 4 convolutional layers with kernel size 3, stride 2, and padding 1 that produces $\mathbf{f}\zero_b, b=0,\dots,3$.

We concatenate $\mathbf{f}\rgb_b$ and $\mathbf{f}\zero_b$ of corresponding spatial resolution along the channel dimension. The resulting four FPV feature maps are fed to a feature pyramid network~\cite{lin2017feature} which produces four tensor maps $\mathbf{H}^{b} \in \mathbb{R}^{256{\times}\textrm{R}_b{\times}\textrm{C}_b}, b=0,\dots,3$, that maintain the same input spatial dimensions and have $256$ channels.

\begin{figure*}[t]  \centering
    \includegraphics[width=1.05
    \linewidth]{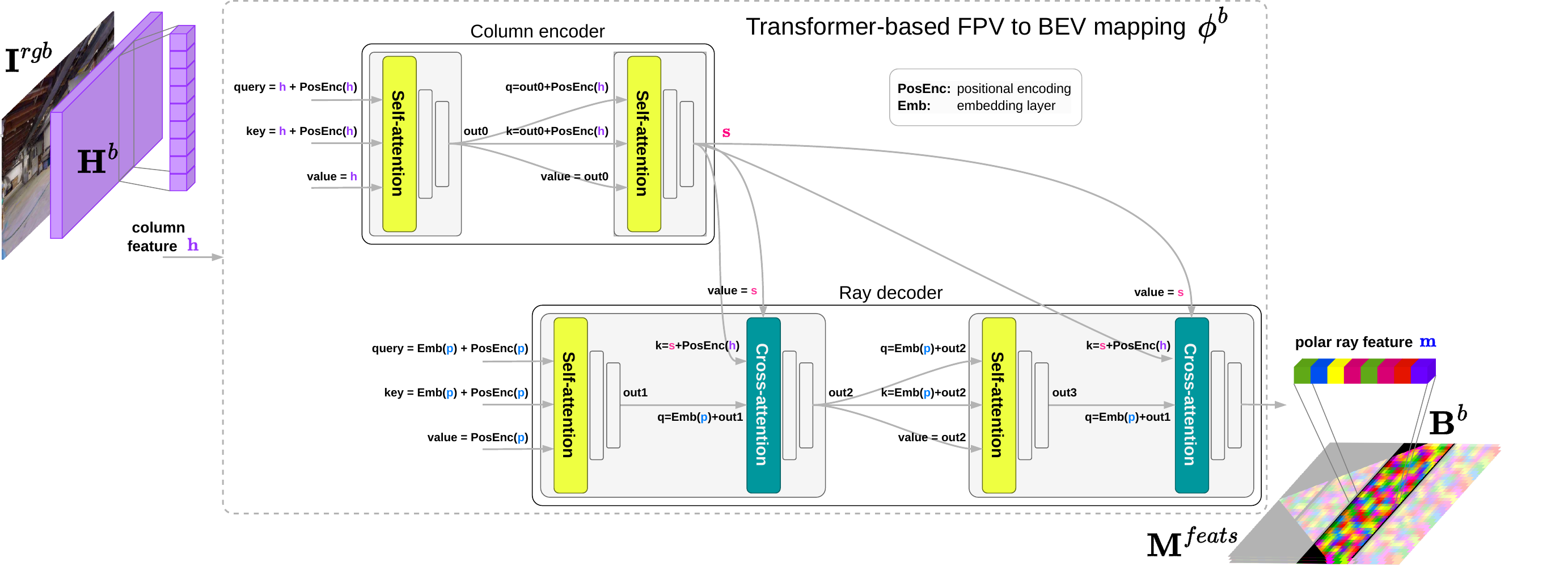}
    \vspace*{-3mm}
    \caption{\label{fig:transformer}\textbf{FPV to BEV transformer architecture} with sequences of self-attention and cross-attention, which query coordinates on the polar ray \textcolor{pos}{$\mathbf{p}$} into the features cells on FPV image columns \textcolor{featcol}{$\mathbf{h}$} to generate polar ray features \textcolor{featray}{$\mathbf{m}$}.
    }
    %\vspace*{-1mm}
\end{figure*}

\myparagraph{FPV to BEV transformers} map the FPV features $\mathbf{H}^{b}$ generated by the feature extractor $\psi$ to the BEV plan, as also formalized in Section~\ref{sec:model}. Each tensor $\mathbf{H}^{b}$ is independently mapped by a network $\phi^{b}$ to a BEV band $\mathbf{B}^{b}$ covering a pre-defined depth range of the BEV --- see Figure~\ref{fig:arch} middle. From now on, for simplicity of notation we drop the band index $^b$. As discussed in Section~\ref{sec:model}, this mapping needs to assign a position in polar coordinates $(\theta, \rho)$ in the BEV band $\mathbf{B}$ for each Cartesian coordinate $(x,y)$ in the FPV tensor $\mathbf{H}$. Following~\cite{saha2022translating}, we exploit the intrinsics of the calibrated camera to identify the correspondence between FPV image feature column $\mathbf{h}$ and BEV polar ray $\theta$ and solve the correspondence problem for each one independently using the transformer-based network $\phi$, which has the structure shown in Figure~\ref{fig:transformer} and comprises two main components: a \textbf{column encoder} on the FPV plane $\mathbf{H}$ and \textbf{a ray decoder} on the BEV band $\mathbf{B}$.
\begin{description}[labelindent=0mm,leftmargin=6mm,topsep=1mm,parsep=0mm,labelsep=0.2em]
    \item[Column encoder] --- computes the self-attention for each column $\mathbf{h} \in \mathbb{R}^{256{\times}\textrm{R}}$ of $\mathbf{H}$ with a standard two-layer \textit{transformer encoder} architecture~\cite{vaswani2017attention} with four attention heads. Each encoder module processes its inputs through the following layers, described with a syntax loosely inspired by PyTorch:
    \begin{enumerate}
        \item \label{MHA1}\texttt{MultiHeadAttention} with 4 heads,
        \item \texttt{Dropout(0.1)} and sum to the Value input of the \texttt{MultiHeadAttention} (layer \ref{MHA1}),
        \item \label{LN1}\texttt{LayerNorm},
        \item \texttt{Linear(in\_dim=256,out\_dim=512)},
        \item \texttt{ReLu} 
        \item \texttt{Droupout(0.1)} and sum to the output of \texttt{LayerNorm} (layer \ref{LN1}),
        \item \texttt{Linear(in\_dim=512,out\_dim=256)},
        \item \texttt{LayerNorm}.
    \end{enumerate}
    As customary, sinusoidal positional encodings are added to the input of each multi-head attention layer. For each column $\mathbf{h}$ we thus obtain at the output a contextualized feature column $\mathbf{s} \in \mathbb{R}^{256{\times}\textrm{R}}$. 
    
    \item[Ray decoder] --- transforms the FPV column feature $\mathbf{s}$ into a BEV ray feature $\mathbf{m} \in \mathbb{R}^{256{\times}\mathrm{P}}$ where $\mathrm{P}$ is the desired ray dimension of the BEV band $\mathbf{B}$. The decoder gets as input the $\mathrm{P}$ target positions in the BEV ray, $\mathbf{p}$, encodes them through a trainable embedding layer and adds to them sinusoidal positional encodings. The mapping is performed by two \textit{transformer decoder} layers, each consisting of a self-attention block and a cross-attention layer between the ray self-attention output and the column feature $\mathbf{s}$. Each transformer decoder has the following layers:
     \begin{enumerate}
        \item \label{R_MHA1}\texttt{MultiHeadAttention1} with 4 heads -- the \textit{polar ray self-attention},
        \item \texttt{Dropout(0.1)} and sum to the Value input of previous \texttt{MultiHeadAttention1} (layer \ref{R_MHA1}),
        \item \texttt{MultiHeadAttention2} with 4 heads -- \textit{cross-attention between polar rays and image columns},
        \item \texttt{Dropout(0.1)} and sum to the output of \texttt{MultiHeadAttention1} (layer \ref{R_MHA1}),
        \item \label{R_LN1}\texttt{LayerNorm1},
        \item \texttt{Linear(in\_dim=256,out\_dim=512)},
        \item \texttt{ReLu}, 
        \item \texttt{Droupout(0.1)} and sum to the output of \texttt{LayerNorm1} (layer \ref{R_LN1}),
        \item \texttt{Linear(in\_dim=512,out\_dim=256)},
        \item \texttt{LayerNorm2}.
    \end{enumerate}   
    This process leads to the BEV ray feature tensor $\mathbf{m} \in \mathbb{R}^{256{\times}\mathrm{P}}$. At this point each FPV image column $\mathbf{h}$ is mapped to a corresponding BEV polar ray $\mathbf{m}$. 
\end{description}
All rays in a BEV band are stacked together along the polar dimension $\theta$ to form the 2D polar feature map $\mathbf{B}$, which is converted to a rectilinear BEV band through an affine transformation. Finally the four bands are stacked together along the ray dimension $\rho$ to obtain the final 2D BEV feature map $\mathbf{M}^{feats} \in \mathbb{R}^{256{\times}100{\times}100}$.

\myparagraph{U-Net/BEV decoder} takes $\mathbf{M}^{feats}$ and outputs $\mathbf{\hat{M}}\zero$ and optionally $\mathbf{\hat{M}}\aux$. Since all processing so far is done on individual rays, the decoder allows to model regularities across rays that the ray-wise transformer cannot capture. The network follows the implementation of Lift-Splat-Shoot~\cite{philion2020lift}, with a convolutional backbone and a structure reminiscent of that of a U-Net. The feature map $\mathbf{M}^{feats}$ is processed by the first three meta-layers of a ResNet18 network~\cite{he2016deep}, leading to three feature maps at different resolutions that are upsampled back to the original spatial resolution of $100{\times}100$, with skip connections between encoder-decoder features of corresponding sizes.

\subsection{Zero-BEV, residual variant}
\label{sec:residual_supp}
The residual variant includes a zero-shot branch (\tcbox[on line,colframe=white,boxsep=0pt,left=0pt,right=0pt,top=1pt,bottom=1pt,colback=orange!15]{in orange} in Figure~\ref{fig:arch}) that uses metric Monocular Depth Estimation (MDE) to generate a geometry-based prediction of the BEV map, $\mathbf{\tilde{M}}\zero$. The core component of the branch is the state-of-the-art Omnidata MDE model~\cite{eftekhar2021omnidata}, which estimates normalized dense depth from RGB images. We took the pretrained \textit{version 2} model~\cite{kar20223d} and finetune it on a dataset of 1M  RGB-metric depth image pairs to generate absolute (i.e. unnormalized) depth --- details will be given below in Section~\ref{sec:mde}.

The estimated depth image is used to compute a 3D point cloud of the scene and \textit{back-project} $\mathbf{I}\zero$ into the 3D camera frame. This is achieved by associating each pixel in $\mathbf{I}\zero$ with the corresponding pixel in the depth image and hence the 3D point cloud. The values in the point cloud are quantized (by counting) into a voxel representation of resolution $100{\times}100{\times}100$. $\mathbf{\tilde{M}}\zero$ is obtained by max pooling the voxel representation to the ground, e.g. summing the voxel content over the height dimension and thresholding values larger than 0. It is important to note that if we consider all height values, voxels would also include points on the ceiling, which are not visible in BEV maps. Therefore we only use height values lower than $2$m. 

Finally, $\mathbf{\tilde{M}}\zero$ is fed to a trainable embedding layer and generates $\mathbf{\check{M}} \in \mathbb{R}^{256{\times}100{\times}100}$, which can be summed to $\mathbf{M}^{feats}$ for the final BEV decoding step.

\subsection{Zero-BEV with inductive bias}

In this model we attempt to introduce an inductive bias to favour disentanglement between geometry and modality. To do that we modify the architecture in Figure~\ref{fig:arch} and the FPV to BEV transformer in Figure~\ref{fig:transformer}. Compared to the base architecture in Figure~\ref{fig:arch}, 
% in this variant and as in previous work~\cite{saha2022translating}, 
it takes the input-output quadruplet including the auxiliary supervision introduced in Section~\ref{ssec:inductivebias} of the main paper and organizes it into two streams, an RGB stream predicting $\mathbf{M}\aux$ from $\mathbf{I}\rgb$ and a zero-shot stream predicting $\mathbf{M}\zero$ from $\mathbf{I}\zero$ and:
\begin{description}
    \item[RGB stream:] $\mathbf{I}\rgb$ is processed by the ResNet50 followed by the feature pyramid network, generating four FPV feature maps. These are fed into the already described transformer, followed by the U-Net decoder.
    \item[Zero-shot stream:]
    $\mathbf{I}\zero$ is still processed by the 4-layer CNN network to generate $\mathbf{f}\zero_b, b=0,\dots,3$, but these are not channel-concatenated with the ones from the RGB stream. Instead they are directly fed into a transformer, followed by a U-Net decoder.
\end{description}
As stated in the main paper, the transformer layers of these two streams are different, but they share the same attention computation, i.e. \textit{inputs and parameters} of Queries and Keys are shared and correspond to those of the RGB stream. This is motivated by the fact that we aim to let attention be identical to the geometric mapping from FPV to BEV, which is calculated from the RGB input image.

The Value projection, however, is not shared: the RGB stream has a classical trainable Value projection, whereas the Zero-shot stream has a Value projection set to the identity function: the output of the transformer is equal to the input (columns from the zero-shot feature map $\mathbf{f}\zero_b$), reweighted by attention. These two streams therefore produce two distinct BEV feature maps, processed by two different BEV decoders (U-Nets), but with shared parameters.

%%%%%%%%%%%%%%%%%%%%%%%%%%%%%%%%%%%%%%%%%%%%%%%%%%%%%%%%%%%%%%%%%%%%%%%%%%%%%%%%%%%%%%%%%%%%%%%%%
%%%%%%%%%%%%%%%%%%%%%%%%%%%%%%%%%%%%%%%%%%%%%%%%%%%%%%%%%%%%%%%%%%%%%%%%%%%%%%%%%%%%%%%%%%%%%%%%%
\section{Finetuning monocular depth estimation}
\label{sec:mde}

Monocular Depth Estimation (MDE) is a central component of the main competing zero-shot baseline and of the residual Zero-BEV model. State-of-the-art MDE foundation models such as MiDaS~\cite{ranftl2020towards} or Omnidata~\cite{eftekhar2021omnidata} reliably estimate dense depth images from RGB input. However, they do that only up to an unknown scaling factor, ie. depth values are normalized within a fixed range, typically $[0,1]$. In this work we require absolute metric depth, and to achieve that we finetune the recent Omnidata \textit{version 2} MDE model, that predicts normalized depth values, with a custom dataset to predict metric depth. The model is a dense prediction transformer (DPT) model based on vision transformers, trained on 10 datasets for a total of approximately 12M image tuples, with 3D data augmentations~\cite{kar20223d} and cross-talk consistency~\cite{zamir2020robust}.  

Additionally, we collect 1M image pairs on the HM3D dataset~\cite{ramakrishnan2021hm3d} consisting of FPV RGB and metric depth images. We create a small validation set of $20,000$ images and use the rest to finetune the MDE model by minimizing the mean absolute difference (MAD) between predicted and ground-truth depth. We train for 20 epochs with batch size 32 and Adam optimizer~\cite{kingma14adam} with learning rate $lr=1e-5$ and weight decay $2e-6$. After optimization, the MDE model achieves a MAD of $13$cm and a relative absolute error of $8\%$ on the validation set.

%%%%%%%%%%%%%%%%%%%%%%%%%%%%%%%%%%%%%%%%%%%%%%%%%%%%%%%%%%%%%%%%%%%%%%%%%%%%%%%%%%%%%%%%%%%%%%%%%
%%%%%%%%%%%%%%%%%%%%%%%%%%%%%%%%%%%%%%%%%%%%%%%%%%%%%%%%%%%%%%%%%%%%%%%%%%%%%%%%%%%%%%%%%%%%%%%%%
\section{Details on the data generation process}

\myparagraph{Cameras/projections to FPV/BEV} the data generation process is an important part of the proposed approach. We use the Habitat simulator~\cite{Savva_2019_ICCV} to render views of the HM3DSem dataset of 3D scenes~\cite{yadav2023habitat}. In order to capture consistent FPV and BEV images, we create two aligned sensors, a FPV pinhole sensor with resolution $384{\times}384$ and field of view (FOV) $79^{\circ}$, and an orthographic camera placed $2$m above the pinhole sensor position, facing down and covering an area of $5\textrm{m}{\times}5\textrm{m}$ in front of the sensor, with resolution of $5$cm per pixel and hence dimensions $100{\times}100$. For each sensor we record RGB, depth and semantic annotations. For FPV, this allows to create $\mathbf{I}\rgb$, the input RGB image, ground-truth semantic segmentation masks used as $\mathbf{I}\zero$ in the zero-shot experiments reported in Table~\ref{tab:results_main} and shown in Figures~\ref{fig:qualitative} and \ref{fig:qualitative_suppmat}, and ground truth depth used for baseline (a.1) in Table~\ref{tab:results_main}. From the orthographic images we obtain ground-truth semantic BEV maps, used \textit{for evaluation only} in the zero-shot experiments and shown in Figures~\ref{fig:qualitative} and \ref{fig:qualitative_suppmat}, and BEV depth, which is thresholded at $10$cm from the floor level to obtain ground-truth navigable and obstacle BEV maps, $\mathbf{{M}}_{nav}\aux$ and $\mathbf{{M}}_{obst}\aux$, optionally used for training. 

All BEV maps, including the zero-shot ones described below, are masked with the FOV of the FPV sensor projected to the BEV map to make reconstruction possible. This mask is computed with a process similar to the one described in Section~\ref{sec:residual_supp}: FPV depth is expanded into a 3D point cloud of the scene, quantized in a voxel representation and projected to the ground by max pooling along the height dimension (again, only considering heights lower than $2$m). Finally the FOV mask is generated by computing the convex hull of this BEV projection.

Concerning the synthetic  \textit{zero-shot data stream} used for training, we explored three different versions: \textit{Synth.}, used for all the results presented in the paper, and two variants, \textit{Mod. Sem.} and \textit{Depth proj.}, discussed in the ablation study summarized in Table~\ref{tab:data_type}. Examples of the three zero-shot data variants for one training image are shown in Figure~\ref{fig:training_data}.

The objectives of the data generation process for the \textit{Synth.} and \textit{Mod. Sem.} variants are similar: generate 2D textures and project them to the 3D scene structure, from where they can be projected to FPV and BEV through perspective and orthographic projections, as described above. This procedure produces a pair of FPV image $\mathbf{I}\zero$ and BEV map $\mathbf{M}\zero$ aligned with the observations captured with the standard process described in the first paragraph of this Section.

\myparagraph{2D $\rightarrow$ 3D mapping} mapping the textures requires a function that assigns 2D positions in the 2D texture files to 3D positions in the scene structure. Given the pseudo-random nature of this function, its exact properties are not important. We therefore piggybacked on the existing projections from the HM3DSem dataset, whose GLB files contain 3D textured meshes: 3D triangle meshes, associated 2D textures, and texture mapping data. From these files, we kept the 3D scene structure and the mapping functions, but replaced the 2D textures, replacing the semantic annotations with pseudo-random data. The difference between this two variants lies in the way the mesh textures are generated. 

\begin{description}
\item[Synth.] --- this is our main variant. We start with blank, e.g. completely black, texture images of the size of the initial semantic texture image present in the original GLB file. We then create random binary structures by randomly selecting texture images from the DTD Dataset \cite{cimpoi14describing} and randomly resizing and thresholding them. We apply morphological opening with a random kernel shape among [\textit{ellipse, square, cross}] and random kernel size $s \in [10,30]$ pixels, and add the resulting structure to the texture image used in the modified GLB file. We repeat this process and keep adding structures until a certain proportion of white pixels is present in each texture image, $5\%$ in our experiments. Finally, we recursively apply morphological dilation with random kernel size $s \in [10,30]$ pixels to the structures present in the texture until we obtain the desired proportion of white matter, from $10\%$ to $30\%$ with intervals of $5\%$ in our experiments. It is worth stressing that this process generates textures whose appearance is decorrelated from the scene geometric structure, as stated in property \textbf{P3}, Section~\ref{ssec:disentangletrain}. The only correlation to the RGB input is up to the scene geometry, which is desired, and arises when the textures are projected on the scene 3D structure.

\begin{figure}[t]  \centering
    \includegraphics[width=1.
    \linewidth]{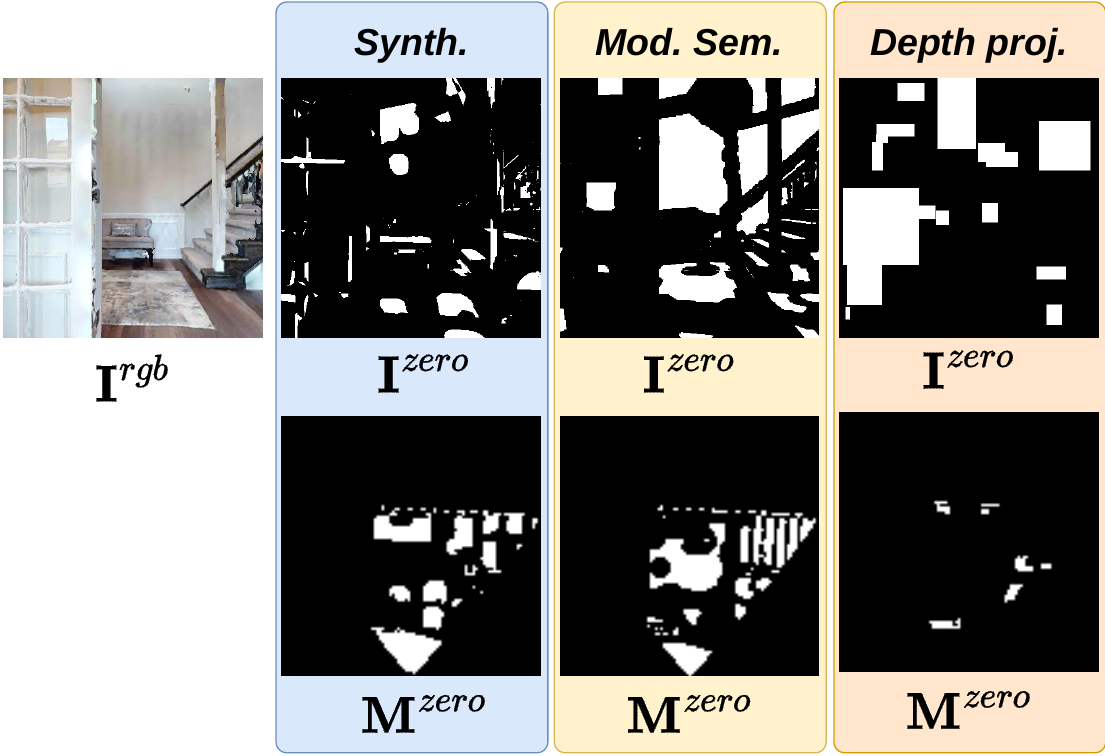}
    \caption{\label{fig:training_data}\textbf{Training Input pairs} for the three different data generation variants evaluated in Table~\ref{tab:data_type}.
    }
    %\vspace*{-1mm}
\end{figure}

\item[Mod. Sem.] --- these data are generated deliberately somewhat violating  property \textbf{P3}, to validate its importance. In this case, we started from the original semantic textures in the original GLB files, thus not discarding them, but binarizing them (so all scene objects become white) and recursively applying morphological erosion with a random kernel shape among [\textit{ellipse, square, cross}] and random kernel size $s \in [10,30]$ pixels, until the desired proportion of white matter is obtained, from $10\%$ to $30\%$ with intervals of $5\%$ in our experiments. In this case, the resulting texture appearance exhibit significantly more correlation with the scene geometry (see Figure~\ref{fig:training_data}). As discussed in Section~\ref{ssec:disentangletrain}, this implies that part of the information needed to reconstruct  $\mathbf{M}\zero$ is already contained in the $\mathbf{I}\rgb$, decreasing the importance of $\mathbf{I}\zero$. This results in lower zero-shot performance, as shown in Table~\ref{tab:data_type}. 

\end{description}

\noindent
\textbf{Depth proj.} --- this third variant uses data generated in a different and somewhat simpler way. In this case we do not modify the scene meshes, instead we start from a synthetic FPV image $\mathbf{I}\zero$ created by adding a random number $n \in [10,20]$ of white rectangles of random size and shape at random positions over a black image. This $\mathbf{I}\zero$ is projected to BEV using ground-truth depth and the same process used to generate  $\mathbf{\tilde{M}}\zero$ and described in Section~\ref{sec:residual_supp}: FPV depth is projected into a 3D point cloud, quantized and projected to the ground by max pooling along the height dimension (only heights lower than $2$m). While $\mathbf{I}\zero$ is now clearly decorrelated from scene structure (see Figure~\ref{fig:training_data}, right), the BEV data generation process relies entirely on FPV depth for the projection to the ground. This implies that the model cannot learn to infer BEV details that are not present in FPV images.

%%%%%%%%%%%%%%%%%%%%%%%%%%%%%%%%%%%%%%%%%%%%%%%%%%%%%%%%%%%%%%%%%%%%%%%%%%%%%%%%%%%%%%%%%%%%%%%%%
%%%%%%%%%%%%%%%%%%%%%%%%%%%%%%%%%%%%%%%%%%%%%%%%%%%%%%%%%%%%%%%%%%%%%%%%%%%%%%%%%%%%%%%%%%%%%%%%%
\section{Losses and Metrics}

\myparagraph{Losses} For training the Zero-BEV models we use the Dice loss~\cite{milletari2016v} averaged over $K$ classes, computed as 
$$\mathcal{L_{D}}^{K} = 1 - \frac{1}{|K|} \sum^{K}_{k=1} \frac{2 \cdot \sum^N_n g^k_n \cdot p^k_n}{\sum^N_n g^k_n + p^k_n + \varepsilon},$$
where $K$ are the number of classes, $N$ are the number of pixels in the BEV maps, $g^{k}_{n}$ is the ground-truth label for pixel $n$ of the BEV map for class $k$, and equals 1 if the pixel belongs to class $k$ and 0 otherwise; similarly $p^{k}_{n}$ is the binary model prediction for pixel $n$, class $k$. Finally, $\varepsilon$ is a small constant to avoid dividing by zero. 

When training Zero-BEV without the auxiliary stream, we only have one class to predict in the zero-shot stream, and the loss is simply expressed as $\mathcal{L_{D}}^{zero} = \mathcal{L_{D}}^{K=1}$. When using the auxiliary data stream we add an auxiliary loss $\mathcal{L_{D}}^{aux}$, which for the best performing model contains $K=2$ classes, navigable and obstacle. The final loss then becomes
$$
\mathcal{L_D} = (1-\alpha) \cdot \mathcal{L_{D}}^{zero} + \alpha \cdot \mathcal{L_{D}}^{aux},
$$
where $\alpha = 0.5$ in all experiments. In the early stages of this work we conducted a sensitivity analysis on the value of $\alpha$ and did not find significant differences.

\myparagraph{Metrics}
Closely related to the Dice loss, zero-shot semantic BEV prediction performance are measured in terms of \textit{Class Intersection over Union (IoU)} and \textit{Pixel IoU}. Class IoU is computed as the ratio between the area of the region where ground-truth and predicted semantic masks for a given class coincide (Intersection), and the area of the region where either of the two predict a detection (Union):  
$$
IoU = \frac{|\textrm{Gr} \cap \textrm{Pr}|}{|\textrm{Gr} \cup \textrm{Pr}|},
$$
where $\textrm{Gr}$ is the ground truth binary semantic BEV mask of the class at hand and $\textrm{Pr}$ is the predicted network output, followed by a sigmoid and thresholded at $0.5$. Please note the slight difference with the Dice loss, which in this notation is expressed as $1 - 2 \cdot |\textrm{Gr} \cap \textrm{Pr}| / |\textrm{Gr}| \cup |\textrm{Pr}|$.

Along the paper we report Average Class IoU, which weights all classes equally. However not all classes have the same frequency (e.g. \textit{floor} and \textit{chair} classes appear much more often than \textit{plant} or \textit{tv}), thus we are also interested in assessing how many pixels are correctly classified overall. For this we compute the Average Pixel IoU, as the cumulated area of all the intersections for all classes and test images, divided by the cumulated area of all unions for all classes and test images.

%%%%%%%%%%%%%%%%%%%%%%%%%%%%%%%%%%%%%%%%%%%%%%%%%%%%%%%%%%%%%%%%%%%%%%%%%%%%%%%%%%%%%%%%%%%%%%%%%
%%%%%%%%%%%%%%%%%%%%%%%%%%%%%%%%%%%%%%%%%%%%%%%%%%%%%%%%%%%%%%%%%%%%%%%%%%%%%%%%%%%%%%%%%%%%%%%%%
\section{Proof of Theorem 3.1 \textit{(Disentangling)}}

\noindent
We consider the column/ray-wise transformer architecture given in Section~\ref{sec:model} of the main paper, and we study how the distribution $\alpha_\rho = \{\alpha_{y,\rho}\}_y$ is calculated through distances between features first-person image features $\mathbf{h}_y$ for a position $y$ and positional encodings $\mathbf{m}_\rho$ on the polar ray, both after trained projections $Q$ and $K$. 

As thought experiment, we will present a possible parametrization of the transformer which encodes a solution to the geometric correspondence problem. Let's start with the (unknown!) \textit{inverse} correspondence mapping from BEV $\mathbf{M}$ to FVP $\mathbf{I}$, which we denote $\gamma(\rho) \rightarrow \{\gamma_{\rho,i}\}_i$ and which maps a position $\rho$ on the polar ray to a set of $y$ coordinates $\gamma_{\rho,i}$ on the corresponding image feature column $\mathbf{h}$. On horizontal or quasi-horizontal surfaces, this mapping produces a single $y$ coordinate, but on vertical walls, a set of multiple coordinates will correspond to a single BEV position. 
%Translating FP images to BEV maps is thus an ill posed problem. 

We now consider the desired zero-shot mapping $\mathbf{h}\rightarrow\mathbf{m}$ which does not change the nature of the observed image, only its geometry, indeed mapping any modality from FPV to BEV. It builds on the geometric mapping $\gamma$, and, as stated in Theorem 3.1, we suppose that it uses average pooling over vertical surfaces, which gives
\begin{equation}
\mathbf{m}_\rho 
%= \textrm{avg}_i \mathbf{h}_{\gamma_{\rho,i}} 
= \frac{1}{N}\sum_i \mathbf{h}_{\gamma_{\rho,i}} 
= \frac{1}{N}\sum_y \mathbf{h}_y
\mathbf{1}_{y\in\gamma_{\rho}} 
\label{eq:gtmapping}
\end{equation}
where $\mathbf{1}_\omega$ is the indicator function giving $1$ if condition $\omega$ is true and $0$ else.
We can see that up to a normalization constant, Eq.~(\ref{eq:gtmapping}) is identical to Eq.~(\ref{eq:attentionoutput}), which gives the attention distribution of the column/ray-wise cross-attention layer. The equality imposes that $W_V=$Id, i.e. the value mapping is the identity. In this case, the indicator function takes the role of the attention distribution $\alpha_{\rho}$, giving
$
\mathbf{m}_\rho=\sum_y \mathbf{h}_y \alpha_{y,\rho} 
$.
Thus, the attention distribution $\alpha_{\rho}{=} \{\alpha_{y,\rho}\}_y$ encodes the belief we have on the correspondence between ray position $\rho$ and all possible column positions $y$. If perfect depth where available, then $\alpha_{\rho}$ were a Dirac distribution with the peak centered on the position correctly aligning the ray position $\rho$ with the input image column.  $\blacksquare$ 

\vspace*{2mm}

\noindent
This ends the proof of Theorem 3.1.

\vspace*{2mm}

\noindent
\myparagraph{Remarks}
It can be seen easily that this cross-attention formulation naturally leads to disentangling: restricting the network to cross-attention alone and minimizing the error on $\mathbf{M}\aux$ will lead the attention distribution to correspond to the geometric transformation $\gamma$, which is used by the second stream to project the zero-shot modality to BEV, without changing it. In particular, zero-shot data properties \textbf{P1} - \textbf{P3} are not required for this solution. We argue that cross-attention is particularly adapted to disentangling geometric transforms and modality translations.
%, in the sense of algorithmic network alignment~\cite{XuWhatCanNNReasonAbout2020}. 

\subsection{Non-linear pooling to the ground} 
The result above holds for any \textit{linear} vertical projection of 3D data to the 2D ground, for instance average pooling or sum pooling. However, most modalities require max-pooling to the ground. For instance, detecting a certain desired object on any height in the image should lead to its appearance on the BEV map regardless of the information above or below of this object.
Adapting the disentangling inductive bias to max-pooling is a non-trivial change, which does not naturally align with transformers.
Attempting to reformulate the mapping to be learned (\ref{eq:gtmapping}), we get
\begin{align}
\label{eq:gtmappingmax}
\mathbf{m}_\rho 
= &\max_i \mathbf{h}_{\gamma_{\rho,i}} \\
= &\max_y \mathbf{h}_y
\mathbf{1}_{y\in\gamma_{\rho}}  \\
\sim= & \sigma_y (\mathbf{h}_y
\mathbf{1}_{y\in\gamma_{\rho}})  
\label{eq:gtmappingsoftmax}
\end{align}
where the last approximation replaces the maximum operator by softmax $\sigma_y$ over items $y$, easier to learn by a neural network. While the individual parts of this function can be easily represented by different modules of a transformer architecture, their specific combination does not easily map into the classical variants. It is easy to see that Eq.~(\ref{eq:gtmappingsoftmax}) does not align with a single layer single head transformer, as the existing softmax is part of the attention distribution representing the uniform ground truth distribution $\mathbf{1}_{y\in\gamma_{\rho}}$. 

One could be tempted to argue that a single attention distribution could actually model the effects of, both, the geometric mapping $\gamma$ and the maximum operation \textit{over elements mapped by $\gamma$}, i.e. a mapping expressed as follows:

\begin{align}
\mathbf{m}_\rho 
= & \max_y \mathbf{h}_y
\mathbf{1}_{y\in\gamma_{\rho}}
\label{eq:gtmappingsoftmax2}
\\
= &\sum_y \mathbf{h}_y
\mathbf{1}_{y\in\gamma_{\rho}}  
\mathbf{1}_{y=\max_{y'} \mathbf{1}_{y'\in\gamma_{\rho}}  }
\label{eq:gtmappingmultimodal}
\end{align}
It can be easily seen, that no set of query and key projections can lead to such a distribution, and that adding multiple heads to a single layer transformer does not change the situation, as the integration over $y$ is done by each head individually.

How about representing (\ref{eq:gtmappingsoftmax}) through multiple layers of multihead attention? If we assume that the geometric mapping $\gamma(\rho) \rightarrow \{\gamma_{\rho,i}\}_i$ is implemented through attention, then we can consider two different cases:
%WORK
\begin{itemize}  
\item[\ding{192}] the selection of the full set  $\gamma_\rho$ of items for each BEV position $\rho$ is done by a single attention distribution. Then the integration over tokens weighted by attention produces a single enriched token, over which no max operation can be performed anymore in the next layers.
\item[\ding{193}] the selection of the full set  $\gamma_\rho$ of items for each BEV position $\rho$ is distributed over multiple attention distributions / heads, each of which would output an enriched vector for a vertical part of the FPV image column. The max operation can then be done by subsequent layers or the feed-forward layers of the self-attention block. However, this representation is hardly practical, as it requires one attention head for each height interval in the scene.
\end{itemize}

\begin{figure*}[t]  \centering
    \includegraphics[width=0.93
    \linewidth]{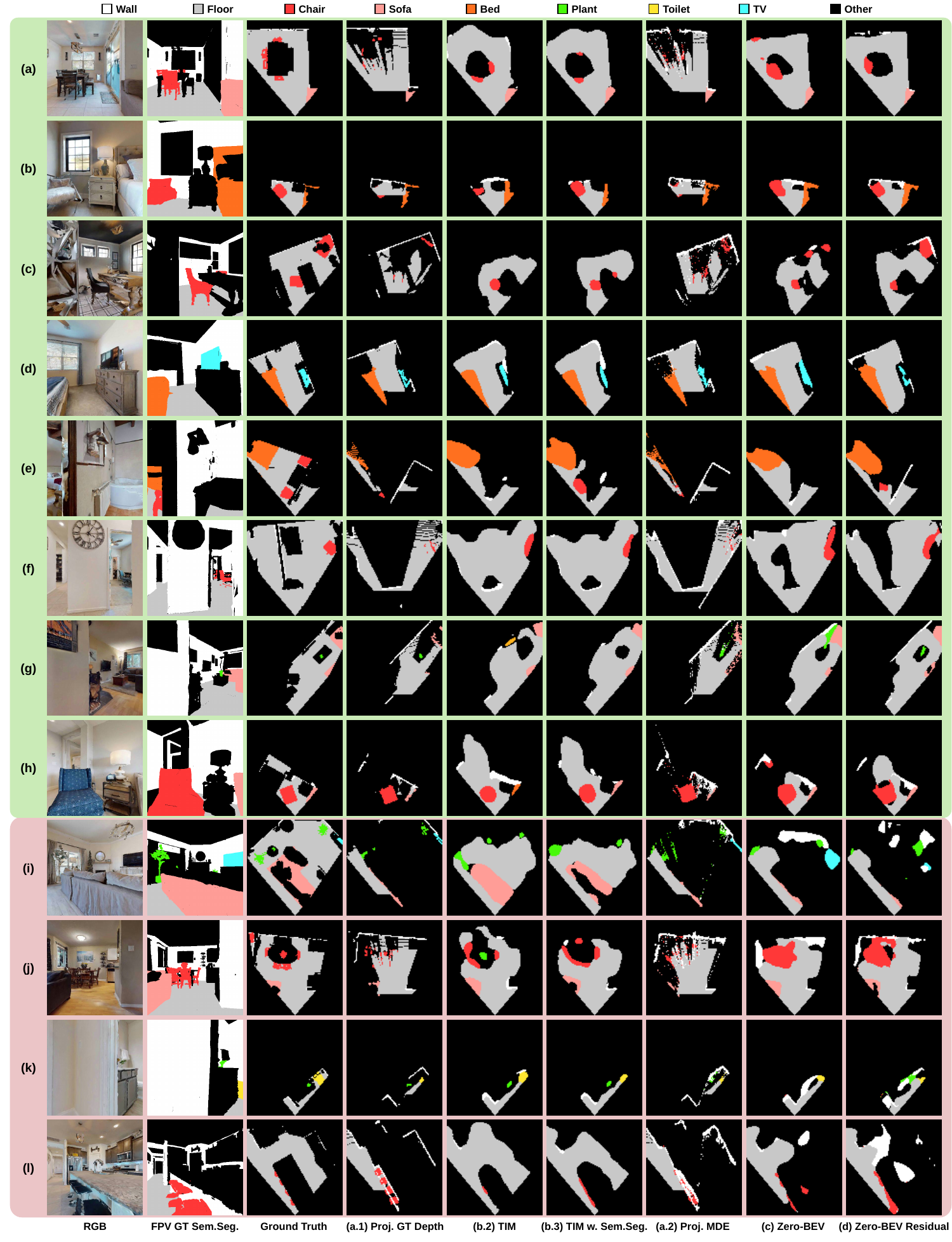}
    \vspace*{-3mm}
    \caption{\label{fig:qualitative_suppmat}\textbf{Qualitative results} on HM3DSem test scenes. (a.1) uses ground-truth depth and methods (c.2) and (c.3) are not zero-shot capable, thus not comparable. (a.2), (c) and (d) are zero-shot models.
    }
    %\vspace*{-1mm}
\end{figure*}

\section{Additional experimental results}

Figure~\ref{fig:qualitative_suppmat} shows additional qualitative results on the zero-shot BEV semantic segmentation task for Zero-BEV variants and baselines. Images in rows (a) to (h) showcase strong performance of our proposed approaches, with results that are clearly superior to the zero-shot alternative (a.2) of back-projecting semantic segmentation masks to the ground using estimated depth. Qualitatively, BEV maps appear very close, if not superior, to those obtained by TIIM, (b.2) and (b.3), a fully-supervised, non zero-shot, state-of-the-art method.

The last four rows of results show cases where Zero-BEV variants achieve relatively poor results. These seem to be cases where input images feature less common perspectives and significant occlusions. Row (j) is interesting because it contains an annotation error: the flower pot on the dining table, on the top part of the BEV map, is wrongly labeled as chair. Standard TIIM (b.2), without access to FPV semantic segmentation masks, correctly predicts a plant on the table, while TIIM (b.3) and Zero-BEV  variants (c) and (d), all using FPV semantic segmentation, struggle to predict reasonable predictions for this part of the scene.

\end{document}